\documentclass[twocolumn]{svjour3}          
\usepackage{graphicx}
\usepackage{multirow}
\usepackage{multicol}
\usepackage{array}
\usepackage{bm}
\usepackage{stfloats}
\usepackage{mathtools}
\usepackage{mathrsfs}
\usepackage{amssymb}
\usepackage{setspace}
\usepackage{algorithm}
\usepackage{pifont}
\usepackage{varwidth}
\usepackage[noend]{algpseudocode}
\usepackage{amsmath}
\usepackage[caption=false]{subfig}
\usepackage{marvosym}

\newcommand{\cmark}{\ding{51}}%
\newcommand{\xmark}{\ding{55}}%

\DeclareMathOperator*{\argmax}{arg\,max}
\DeclareMathOperator*{\argmin}{arg\,min}

\usepackage{ae}
\usepackage{aecompl}

\begin{document}

\title{Deep Bayesian Self-Training
}

\titlerunning{Deep Bayesian Self-Training}

\author{Fabio~De~Sousa~Ribeiro \and Francesco~Caliv\'a \and Mark~Swainson \and Kjartan~Gudmundsson \and Georgios~Leontidis \and Stefanos~Kollias
}
\institute{\Letter \ Fabio De Sousa Ribeiro, Francesco Caliv\'a, \\Georgios Leontidis, and Stefanos Kollias \at
              Machine Learning Group \\
              University of Lincoln, UK \\
              \email{\{fdesousaribeiro, fcaliva,\\gleontidis, skollias\}@lincoln.ac.uk}
           \and
           Mark Swainson, Kjartan Gudmundsson \at
              National Centre for Food Manufacturing \\ 
              Holbeach Technology Park, UK \\
              \email{\{mswainson, kgudmundsson\}@lincoln.ac.uk}
              }
\date{ \vspace{-1em} }
\maketitle
\begin{abstract}
Supervised Deep Learning has been highly successful in recent years, achieving state-of-the-art results in most tasks. However, with the ongoing uptake of such methods in industrial applications, the requirement for large amounts of annotated data is often a challenge. In most real world problems, manual annotation is practically intractable due to time/labour constraints, thus the development of automated and adaptive data annotation systems is highly sought after. In this paper, we propose both a (i) Deep Bayesian Self-Training\footnote{Code available at: {https://github.com/fabio-deep/Deep-Bayesian-Self-Training}} methodology for automatic data annotation, by leveraging predictive uncertainty estimates using variational inference and modern Neural Network (NN) architectures, as well as (ii) a practical adaptation procedure for handling high label variability between different dataset distributions through clustering of NN latent variable representations. An experimental study on both public and private datasets is presented illustrating the superior performance of the proposed approach over standard Self-Training baselines, highlighting the importance of predictive uncertainty estimates in safety-critical domains.
\end{abstract}
\section{Introduction}
\label{sec: intro}
With the advent of Big Data in industrial applications, the ability to automatically label datasets using limited supervision is increasingly sought after. In most real world problems, manual annotation is practically intractable due to time and labour constraints. Furthermore, recent advances in Supervised Deep Learning have shown that training over parameterised models on large datasets significantly increases performance~\cite{krizhevsky2012imagenet}. With that in mind - and despite the high demand for annotated data - deep learning practitioners have not yet explored or leveraged many of deep learning tools for automatic annotation systems. This is evidenced by the scarcity of existing research in the field, compared to related others~\cite{gal2017deep}. Automated annotation techniques typically involve semi-supervised algorithmic variants, wherein learning systems are often trained on a small initial sample of labelled data, and leverage information from unlabelled data to generalise better~\cite{zhu2006semi}. Well established semi-supervised methods such as Self-Training~\cite{yarowsky1995unsupervised}, Transfer Learning~\cite{pratt1993discriminability}, Co-Training~\cite{blum1998combining}, Active Learning~\cite{cohn1996active} and Tri-Training~\cite{zhou2005tri} among others have shown to be useful for labelling in the past, but some challenges remain with regards to their scalability to high dimensional data and their suitability to modern Deep Learning settings~\cite{tong2001active,gal2017deep}. 
Prominent recent works have explored some of these ideas in the context of modern deep models, proposing new paradigms such as Co-teaching~\cite{han2018co}, Active Learning on Image data~\cite{gal2017deep} and analysing Deep Transfer Learning~\cite{yosinski2014transferable,zamir2018taskonomy} with good levels of success. 
Taking inspiration from these works, in this paper we primarily focus on exploring the Self-Training algorithm in combination with modern Bayesian Deep Learning methods, and leverage predictive uncertainty estimates for self-labelling of high dimensional data.  
\subsection{\textbf{Background on Application Domain}}
In addition to public domain datasets, we evaluate our methods on a real world task involving Optical Character Verification (OCV) of real food packaging images, expanding on earlier work in~\cite{de2018adaptable} by reducing manual data annotation. 

Incorrectly labelled food products (e.g. bearing an incorrect/illegible \textit{use-by} date) result in product recalls and food waste, as label faults can lead to food safety incidents. Label faults are primarily attributed to human error during error-prone manual checking. Automatic approaches typically involve OCV, whereby a supervisory system holds the correct date code string and transfers it to both the printer and the vision system. The latter will then verify its read, and take appropriate action. Such a system could also be used alongside other systems, such as blockchain, within the food chain for food traceability~\cite{pearson2019distributed}. Current OCV systems require accurately labelled data to be utilised for training, but the labelling process is time consuming, expensive and requires expertise. They also rely on consistency in date code format, packaging and camera view angle which is difficult to ensure in a manufacturing environment, so there is a great need for a more robust solution. 
\subsection{\textbf{Contribution}}
We propose a Deep Bayesian Self-Training methodology orthogonal to~\cite{gal2017deep}, that leverages approximate variational inference in DNNs to estimate predictive uncertainty during a Self-Training setting. Both aleatoric and epistemic uncertainties of predicted pseudo-labels for unseen data are estimated, and the samples with the lowest predictive uncertainty (highest confidence) are added to the training set in an automated manner. We offer ways to mitigate the known problem of propagating errors in Self-Training by including: (i) an entropy penalty on the log likelihood loss to punish over confident output distributions and facilitate thresholding, and (ii) an adaptive sample-wise weight on the influence of predicted pseudo-labelled samples over gradient updates to be inversely proportional to their predictive uncertainty. Lastly, we propose a new simple methodology for visualising and analysing variability between two dataset distributions in DNNs, and attempt to adapt information from one problem to the other by clustering learnt latent variable representations in the context of our application domain. An experimental study on both public and private (real) datasets is presented demonstrating the increased performance of our algorithm over standard Self-Training baselines.
\section{Related Work}
\label{sec: related work}
Deep Learning model's ability to learn abstract hierarchical representations from data has pushed the state-of-the-art in most machine learning related tasks~\cite{krizhevsky2012imagenet,huang2017densely}. The uptake of these methodologies in academia and industry has resulted in many diverse and interesting DNN applications, wherein patterns learned from data have been adapted to perform tasks in various domains, including Computer Vision~\cite{ribeiro2018end,huang2017densely,sun2017revisiting,de2018adaptable}, Medical Imaging~\cite{chudzik2018microaneurysm,kollias2017adaptation,havaei2017brain} and Signal Processing~\cite{caliva2018deep,ribeiro2018towards}. Although many important improvements to DNNs have been made in various domains, there are still many adversities in training models which can be easily adapted to other tasks; and the lack of annotated data is one of the contributing factors. 
\subsection{\textbf{Deep Semi-Supervised Learning}}
Most related work addressing the aforementioned issues is often related to domain adaptation philosophy and semi-supervised learning algorithms such as: Self-Training~\cite{yarowsky1995unsupervised}, which is an iterative procedure for self-labelling data points in an unlabelled pool, and retraining a classifier until stop conditions are met. Co-Training~\cite{blum1998combining}, can be considered multi-view variant of Self-Training wherein two separate classifiers are trained on different views of the data, and augment each others training sets with their predicted labels. Tri-Training~\cite{zhou2005tri} extends Co-Training by having three classifiers, and unlabelled examples are added to a classifier's training set iff. the other two agree on the predicted label. Active Learning~\cite{cohn1996active} selects the most informative samples from a pool of unlabelled data, and retrains the classifier with human given labels in an effort to maximise performance and minimise data labelling requirements. Transfer Learning~\cite{pratt1993discriminability} is often used when there is a lack of annotated data in the target domain, and the goal is to adapt knowledge from one task to another by initialising the weights of the target task with the pre-trained weights of another, often performing better than random initialisation. Among these algorithms, Transfer Learning has undoubtedly had the most success in the context of deep models, and it is widely used in computer vision for adapting visual features from large source domains, to target domains with limited annotated data. Notably, \cite{yosinski2014transferable} find that; initializing a network with transferred features boosts generalization that lingers even after fine-tuning to the target dataset, and transferring features from distant tasks is still better than using random weights. Recent work in~\cite{kaiser2017one} suggests that a single DL model can jointly learn a number of tasks from multiple domains successfully. In fact, it was observed that adding knowledge from unrelated tasks never hurts performance, rather mostly improves it on all tasks. This phenomenon is complimented by research in~\cite{doersch2017multi}, with results suggesting that combining tasks, even via a na\"ive multihead architecture, always improves performance. Authors in~\cite{zoph2017learning} propose learning a network comprised of the most successful layers from many different source networks, which are continuously generated and evaluated by a Recurrent Neural Network (RNN) controller. Task Transfer Learning was recently studied in great depth by~\cite{zamir2018taskonomy}, where a fully computational approach termed Taskonomy was proposed. This was achieved by identifying dependencies between $26$ different tasks in latent space, producing a computational taxonomic map for task Transfer Learning. Deep generative modelling is also gaining popularity in tackling adaptation of knowledge learnt from data generating distributions to pool sets of unlabelled data~\cite{kingma2014semi,tzeng2017adversarial,bousmalis2017unsupervised}. Other notable related works presented more recently include Co-Teaching~\cite{han2018co}, wherein two neural networks are trained simultaneously and teach each other to select clean labels, then decide what data to use for training. Mean teacher models~\cite{tarvainen2017mean} maintain an exponential moving
average of model weights and penalise inconsistent predictions, enabling training with fewer labels as an added benefit. Deep Co-Training~\cite{qiao2018deep} extends the original Co-Training algorithm by training multiple DNNs with different views generated by exploiting adversarial examples. In~\cite{lee2013pseudo} a simple method termed Pseudo-Label similar to Entropy Regularisation~\cite{grandvalet2006entropy} is proposed, and it consists of iteratively assigning pseudo-labels via the maximum predicted probability of a NN. Although research on Self-Training with deep models is scarce, notable work in~\cite{zou2018unsupervised} presents an unsupervised domain adaptation (UDA) framework based on Self-Training for semantic segmentation using DNNs. They develop a self-paced policy that increases the number of pseudo-labels incorporated in each additional round, and demonstrate performance benefits over other popular methods. However, as is the case with all previous works mentioned thus far, their proposed approach does not provide principled predictive uncertainty estimates. The black box nature of DNNs is a concern in most real world applications, and by quantifying what a model doesn't know with uncertainty measures, we can not only better trust our predictions but also avoid potentially harmful outcomes~\cite{kendall2017uncertainties}. With that in mind, perhaps the most significant related work is in~\cite{gal2017deep}, where the authors propose a Bayesian formulation of Active learning for image data using DNNs, obtaining a significant improvement on existing active learning approaches by considering uncertainty estimates in approximating acquisition functions.
\subsection{\textbf{Uncertainty Estimation}}
The estimation of uncertainty as a measure of confidence over a model's predictions is desireable for self-labelling, and for safety-critical systems in general~\cite{kendall2017uncertainties}. Bayesian Neural Networks (BNNs) were studied by many in the past~\cite{denker1991transforming,neal2012bayesian,mackay1992practical} and have more recently regained popularity. In BNNs, uncertainty is typically captured by placing a prior distribution, such as a Gaussian, over the weights and averaging over all possible parameters, rather than optimising them directly. Bayesian inference is then used to compute the posterior over the weights capturing the set of likely parameters. However, BNNs are difficult to perform inference in with traditional methods, as they do not scale well scale to high dimensional inputs or very complex DL models~\cite{kendall2017uncertainties}. Recent promising methods including~\cite{kendall2017uncertainties,gal2016uncertainty,gal2015bayesian} offer alternative ways of capturing uncertainty by simple modifications to loss functions, and having the network learn/predict aleatoric uncertainty in an unsupervised manner. Aleatoric uncertainty relates to sensory noise in the acquisition process of the data, and is therefore inherently irreducible~\cite{kendall2017multi}. However, we argue that it can be a great tool for quantifying our uncertainty about pseudo-label predictions. In~\cite{gal2015bayesian}, Dropout was shown to perform approximate variational inference, wherein stochastic forward passes with Dropout at test time are effectively samples from the approximate posterior. This technique is know as Monte Carlo (MC) Dropout~\cite{gal2015bayesian}, and can be used to quantify epistemic uncertainty in NN predictions. Epistemic uncertainty relates to our uncertainty about the model parameters, which is in fact reducible as we observe more data. This is because we can explain the uncertainties about the model parameters in the limit of observing all explanatory variables of the data~\cite{kendall2017multi,kendall2017uncertainties}. This type of uncertainty is useful for identifying out-of-distribution data points, and is the most important type of uncertainty measure when assigning pseudo-labels to data.

In this paper we argue that, with some modifications, uncertainty estimation techniques in Bayesian deep learning can also be useful in a Self-Training setting, and to the best of our knowledge these ideas have yet to be explored in this context. All things considered, we propose a Deep Bayesian Self-Training algorithm, in which a DNN assigns pseudo-labels to new data, and automatically weighs their sample-wise importance for the next Self-Training iteration to be inversely proportional to the predictive uncertainly of the assigned pseudo-label. In this way, we can reduce the burden of manual data annotation requirements, and also offer a measure of uncertainty about our predictions which is important in safety-critical domains.

\section{\textbf{Deep Bayesian Self-Training}}
In this section, we provide a brief background on Bayesian NNs, and explore the idea of uncertainty estimation of pseudo-label predictions for unlabelled data, in a Deep Bayesian Self-Training framework (see Algorithm 1).
In order to quantify what our and does not know, we extend existing approaches for estimating uncertainty in deep CNNs~\cite{kendall2017uncertainties,gal2016dropout}. To this end, we consider the following Bayesian formulation of a deep CNN for estimating both aleatoric and epistemic uncertainties. 
%
\subsection{\textbf{Bayesian Neural Networks}}
Let $\mathcal{D} = \{(\mathbf{X}, \mathbf{Y})\}$ denote a dataset given as $N$ pairs of inputs $\mathbf{x}_i\in \mathbb{R}^d$ of dimension $d$, and class labels $\bm{y}_i \in \{1, \dots K\}$ of $K$ total classes. Assuming a Bayesian Neural Network (BNN) formulation, we place a Gaussian prior probability distribution $p(\bm{\omega})$ over the set of trainable parameters $\bm{\omega} = \{\mathbf{W}_1,\dots,\mathbf{W}_\ell\}$. We define the likelihood conditional output distribution $p(\mathbf{Y}|\mathbf{X}, \bm{\omega})$ of NN for mapping inputs to labels, by finding parameters $\bm{\omega}$ that yield the Maximum Likelihood Estimate (MLE). MLE is the pillar of supervised learning in DNNs and is defined as
\begin{equation}
\bm{\widehat{\omega}}_{\text{ML}} =  \argmax_{\bm{\omega}} \ \sum_{i=1}^{N} \log p(\bm{y}_i|\mathbf{x}_i, \bm{\omega}),
\end{equation}
yielding a point estimate for the most likely parameters to have generated the data. In a Bayesian sense, the MLE is a special case of Maximum A Posteriori (MAP) estimation when a uniform prior is assumed. In practical classification tasks, the MLE estimator is obtained by minimising the negative log-likelihood of a Bernoulli or softmax distribution depending on the number of classes. We define the softmax negative log-likelihood of our classification NN model as
\begin{equation}
-\log p(\bm{y}_i = k|\mathbf{x}, \bm{\omega}) = -\bigg(\mathbf{z}_k - \log \sum_{k'}\exp(\mathbf{z}_{k'})\bigg)
\end{equation}
where $\mathbf{z}$ denotes the vector of output logits by the network and $k$ denotes a class. Having defined a prior and a likelihood, we would like to compute the posterior probability distribution over the weights given the data by Bayes rule 
\begin{equation}
    p(\bm{\omega}|\mathbf{X}, \mathbf{Y}) = \frac{p(\mathbf{Y}|\mathbf{X}, \bm{\omega})p(\bm{\omega})}{p(\mathbf{Y}|\mathbf{X})}\propto p(\mathbf{Y}|\mathbf{X}, \bm{\omega})p(\bm{\omega}),
\end{equation}
with which we can also formulate the predictive distribution given new inputs $\mathbf{x}^{*}$ and labels $\bm{y}^{*}$
\begin{equation}
p(\bm{y}^{*}|\mathbf{x}^{*}, \mathbf{X}, \mathbf{Y}) = \int p(\bm{y}^{*}|\mathbf{x}^{*}, \bm{\omega})p(\bm{\omega}|\mathbf{X}, \mathbf{Y})\text{d}\bm{\omega},
\end{equation}
enabling predictions using a full distribution over the parameters $\bm{\omega}$, which captures uncertainty over the model parameters, rather than using a point estimate. However, in most cases, the posterior distribution $p(\bm{\omega}|\mathbf{X}, \mathbf{Y})$ cannot be evaluated analytically. This is because to compute the marginal probability $p(\mathbf{Y}|\mathbf{X})$ we must integrate over all possible model parameters $\bm{\omega}$ with weighted probability $p(\bm{\omega})$, in order to obtain the normalising constant, also known as the model evidence. Since the true posterior distribution $p(\bm{\omega}|\mathbf{X}, \mathbf{Y})$ is intractable, various approximations exist~\cite{mackay1992practical,hinton1993keeping,neal2012bayesian}. Most of them were important early steps towards performing approximate inference in Bayesian NNs, but are unfortunately difficult to employ in modern applications due to scalability constraints or expert knowledge requirements. More recent work in~\cite{graves2011practical,welling2011bayesian,gal2016dropout,kingma2013auto} addressed some of these issues with variational inference, reigniting interest in the field of Bayesian NNs.
\subsection{\textbf{Variational Inference}}
Next, we provide a background on variational inference (VI) to contextualise some of the ideas presented in~\cite{gal2016dropout}, wherein Dropout is shown to perform approximate variational inference in NNs when used at test time. In VI, a factorised variational distribution from a tractable family $q_{\theta}(\bm{{\omega}})$, parameterised by $\theta$, is defined for approximating the posterior distribution by minimising the Kullback-Leibler (KL) divergence between $q_{\theta}(\bm{{\omega}})$ and $p(\bm{\omega}|\mathbf{X}, \mathbf{Y})$. Intuitively, the KL divergence is a non-negative asymmetric measure of similarity between the two distributions $\text{KL}(q_{\theta}(\bm{{\omega}}) \ || \ p(\bm{\omega}|\mathbf{X}, \mathbf{Y}))$,
which we minimise via the variational parameters $\theta$ of our approximating distribution $q_{\theta}(\bm{{\omega}})$
\begin{equation}
\widehat{\theta} =\argmin_{\theta} \ \mathbb{E}_{q_{\theta}(\bm{{\omega}})}
\big[\log q_{\theta}(\bm{{\omega}}) - \log p(\bm{\omega}|\mathbf{X}, \mathbf{Y})\big].
\end{equation}
However, optimising the KL divergence directly requires knowledge of the intractable posterior. This is circumvented by instead maximising the evidence lower bound (ELBO) on the marginal log-likelihood $\log p(\mathbf{Y}|\mathbf{X})$, derived via Jensen's inequality $\log(\mathbb{E}[X]) \geq \mathbb{E}[\log(X)]$
\begin{equation}
\mathcal{L}_{\text{ELBO}}(\theta) =\log p(\mathbf{Y}|\mathbf{X})  -\text{KL}(q_{\theta}(\bm{{\omega}}) \ || \  p(\bm{\omega}|\mathbf{X}, \mathbf{Y})),
\end{equation}
and given that the KL divergence $\geq 0$ then
\begin{equation}
\log p(\mathbf{Y}|\mathbf{X}) = \mathcal{L}_{\text{ELBO}}(\theta) + \text{KL}(q_{\theta}(\bm{{\omega}}) \ || \  p(\bm{\omega}|\mathbf{X}, \mathbf{Y})).
\end{equation}
By maximising the lower bound we implicitly maximise $\log p(\mathbf{Y}|\mathbf{X})$, and minimise the KL divergence as intended. 
We extend these ideas in light of recent developments in~\cite{gal2016dropout} with the Monte Carlo Dropout approximation using $q_{\theta}(\bm{{\omega}})$, further explained in the following section. 
%
\begin{figure*}
{\centering
\begin{minipage}{.85\textwidth}
\begin{algorithm}[H]
\centering
\caption{Deep Bayesian Self-Training}
\begin{spacing}{1.3}
\begin{algorithmic}[1]
\State {\textbf{Note:} Pseudo code for training a progressively growing DenseNet in a Bayesian Self-Training setting. The incremental growth factor $\nu$ is adjusted for dataset size.}
\vspace{1mm}
\hrule
\vspace{1mm}
\Function{DBST}{$\mathcal{D} = \{(\mathbf{x}, \bm{y}, \lambda)\}_{i=1}^N \ ,  \ \mathcal{U}=\{\widetilde{\mathbf{x}}\}_{i=1}^{\widetilde{N}}$}\Comment{Input training and unlabelled datasets}
\State {Initialise :}{$\ r \gets 0$, $k \gets 12$, $k_{\mathrm{max}} \gets 24$}
\State {Initialise}{$\ \forall \ \mathbf{x}_i \in \mathcal{D}$ : $\lambda_i \gets 1$}
\While{$|\mathcal{U}| > 0$}\Comment{Unlabelled dataset cardinality}
\State {$r \gets r+1$, $k \gets \min(k + \nu \cdot (r - 1), k_{\mathrm{max}})$}
\State {$f(\mathbf{x}) \gets$} \Call{train}{$\mathcal{D}, k$}\Comment{Train a DenseNet with growth rate $k$}
\State {$(\widehat{\mathbf{p}},\widehat{\mathbf{s}}) \gets$} \Call{MC Dropout}{$f(\mathbf{x}),\mathcal{U}$}

\For{$\widetilde{\mathbf{x}}_i \in \mathcal{U}$}
\State {$\mathrm{Var}[ \bm{y}_i] \gets \exp(\widehat{\mathbf{s}}_i) + \mathbb{H}\big[\widehat{\mathbf{p}}_i\big]$}\Comment{Aleatoric and Epistemic uncertainties}
\State {$\widehat{\bm{y}}_i \gets \argmax \widehat{\mathbf{p}}_i$}
\If{$\mathrm{Var}[\bm{y}_i] < \tau $}\Comment{$\tau$ is computed via IQR}
\State {$\lambda_i \gets 1 / \exp({\mathrm{Var}[\bm{y}_i]})^{\phi(r)}$}
\State {$\mathcal{D} \gets \mathcal{D} \cup \{\widetilde{\mathbf{x}}_i,  \widehat{\bm{y}}_i, \lambda_i\}$}\Comment{Add weighted pseudo-labelled sample to $\mathcal{D}$}
\EndIf
\EndFor
\EndWhile
\vspace{.7mm}
\EndFunction
\end{algorithmic}
\end{spacing}
\end{algorithm}
\end{minipage}
\par}
\end{figure*}
\subsection{\textbf{Continuous Relaxation of Dropout}}
Concrete Dropout is based on concrete relaxation of discrete distributions~\cite{maddison2016concrete}, allowing the replacement of Dropout's discrete Bernoulli distribution with its continuous relaxation~\cite{gal2017concrete}.
To obtain calibrated uncertainty estimates with Monte Carlo Dropout, it is necessary to tune the Dropout probabilities. A grid-search is a common but costly approach for large models, highlighting the benefit of optimising them directly with Gradient Descent. This requires formulating an objective for minimising epistemic uncertainty~\cite{gal2016dropout} using the variational interpretation of Dropout. 

Formally, Dropout can be treated as an approximating distribution $q_{\theta}(\bm{\omega})$ to the posterior in a BNN, where $\bm{\omega}$ represents the weight matrices of the $\ell^\text{th}$ of $L$ layers in the network $\bm{\omega} = \{\mathbf{W}_{\ell}\}^{L}_{\ell=1}$, and $\theta$ are the variational parameters to optimise~\cite{gal2017concrete}. Let $\mathcal{F}(\bm{\omega})$ be the model with weight matrix realisation $\bm{\omega}$; given a random set $S$ comprising $M$ of all $N$ data points, and denote the model's output on the $\mathbf{x}_{i}$ input as $\mathcal{F}(\mathbf{x}_{i};{\bm{\omega}})$. The following NN objective function can then be formulated
\begin{multline}
    \mathcal{\hat{L}}_{\text{MC}}(\theta) = -\frac{1}{M}\sum_{i\in S} \log p(\bm{y}_{i}| \mathcal{F}(\mathbf{x}_{i};{\bm{\omega}})) \ +\\
    \frac{1}{N} \mathrm{KL}(q_{\theta}(\bm{\omega}) \ || \ p(\bm{\omega})),
\end{multline}
where $p(\bm{y}_{i}|\mathcal{F}(\mathbf{x}_{i};{\bm{\omega}}))$ is the model's likelihood, a Gaussian with a predictive mean given by $\mathcal{F}(\mathbf{x}_{i};{\bm{\omega}})$. KL is a regularisation term which constrains the approximate posterior $q_{\theta}(\bm{\omega})$ from deviating too far from prior $p(\bm{\omega})$. Following~\cite{gal2016uncertainty} we can approximate the $\text{KL}$ term with
\begin{equation}
\text{KL}(q_{M}(\mathbf{W}) \ || \ p(\mathbf{W}))\propto \frac{l^{2}(1-p)}{2}||\mathbf{M}||^{2}-K\mathbb{H}[p],
\end{equation}
where $\{\mathbf{M}_{\ell},p_{\ell}\}_{\ell=1}^{L}$ is a set of mean weight matrices and Dropout probabilities, such that (s.t.) $q_{M_{\ell}}(\mathbf{W}_{\ell}) = \mathbf{M}_{\ell}\cdot \text{diag}[\text{Bernoulli}(1-p_{\ell})^{K_\ell}]$ for a single NN weight matrix $\mathbf{W}_{\ell}\in\mathbb{R}^{K_{\ell+1}\times K_{\ell}}$. $\mathbb{H}[p]$ is simply the entropy of a Bernoulli random variable with probability $p$
\begin{equation}
    \mathbb{H}[p]\coloneqq -p\log p-(1-p)\log(1-p),
\end{equation}
which can be interpreted as a regularisation term that only depends on Dropout probability $p$, so minimising the KL term is equivalent to maximising the entropy of a Bernoulli random variable with probability $(1-p)$. Rather than sampling the random variable from the discrete Bernoulli distribution, by adopting the Concrete distribution~\cite{maddison2016concrete,gal2017concrete} with some temperature $t$, it is possible to sample variables in the interval $[0,1]$, s.t. the concrete relaxation distribution $\widetilde{\mathbf{z}}$ 
\begin{multline}
\label{eq: concrete relaxation of the dropout mask}
\widetilde{\mathbf{z}}=\text{sigmoid}\biggl(\frac{1}{t}\cdot\big[\log p - \log(1-p) \ + \\ \log u - \log(1-u)\big] \biggr),
\end{multline}
parametrised by means of $u\sim\text{Unif}(0,1)$, provides a relationship between $\widetilde{\mathbf{z}}$ and $u$, which is differentiable w.r.t. $p$. With the concrete relaxation of the dropout masks, the Dropout probabilities for each layer  $\{p_\ell\}_{\ell=1}^{L}$ can be optimised using the pathwise derivative estimator~\cite{gal2017concrete}.
\subsection{\textbf{Entropy Penalty on Output Distributions}}
The probabilities assigned to incorrect classes at test time help quantify a model's ability to generalise. By penalising output distributions with low entropy (i.e. confident predictions), we can obtain a similar effect to label smoothing and improve generalisation~\cite{pereyra2017regularizing}. This can useful in Self-Training, since we assign pseudo labels-based on low uncertainty predictions, which are in some cases wrongly assigned. We suggest that by penalising very confident output distributions we can improve generalisation, and make thresholding easier since the output distributions are smoother, rather than overly concentrated at $0$ or $1$. The entropy of a NNs output conditional distribution is given by
\begin{equation}
    \mathbb{H}\big[p(\bm{y}|\mathbf{x},\bm{\omega})\big]=-\sum_i p(\bm{y}_i|\mathbf{x}, \bm{\omega}) \log p(\bm{y}_i|\mathbf{x},\bm{\omega}),
\end{equation}
with $p(\bm{y}|\mathbf{x}, \bm{\omega})$ as the probability distribution obtained from a softmax function. To penalise very confident predictions we can simply take the negative log-likelihood, and subtract the entropy of the output distribution as
\begin{equation}
    \mathcal{L}_{\mathrm{NLL}}(\bm{\omega})=-\sum \log p(\bm{y}|\mathbf{x}, \bm{\omega})-\beta \mathbb{H}\big[p(\bm{y}|\mathbf{x}, \bm{\omega})\big],
\end{equation}
where the scaling hyperparameter $\beta$ balances how much we'd like to penalise non-uniformity of the softmax.
\subsection{\textbf{Inverse Uncertainty Weighting}}
\label{subsec: inverse uncertainty weighting}
A known limitation of Self-Training is the potential accumulation of wrongly pseudo-labelled samples being added to the training set. A common approach is to remove less confident samples from the training set, and leave them in the unlabelled set. However, this tends to under perform in practice, as the algorithm can become biased by continuously adding the easiest unlabelled samples to the training set. This can hinder learning over time, as more difficult and potentially informative samples are neglected.

In attempt to mitigate this behaviour, we propose a sample-wise weighting scheme during training that places a weight on each training sample $\{\mathbf{x}_i, \widehat{\bm{y}}_i, \lambda_i\}$, proportional to the predictive uncertainty over its pseudo-label $\widehat{\bm{y}}_i$, such that its contribution to the loss function is inversely proportional to its predictive uncertainty (see Algorithm 1). To calculate the predictive uncertainty we can have the network predict the aleatoric uncertainty as one of its outputs, and add the epistemic uncertainty obtained from the variance of Monte Carlo Dropout samples. 

Formally, let $\widehat{\mathbf{p}}_{t}=\mathrm{softmax}(\mathcal{F}(\mathbf{x};\bm{\widehat{\omega}}_t))$ denote the softmax out of a BNN, and $\{\widehat{\mathbf{p}}\}_{t=1}^{T}$ be the set of outputs from $T$ Monte Carlo Dropout samples at test time, each parameterised by weights drawn from the approximate posterior $\bm{\widehat{\omega}}_t \sim q_{\hat{\theta}}(\bm{\omega})$. We propose calculating the predictive uncertainty from these samples by generalising the binary variant approach in~\cite{kwon2018uncertainty} to a multivariate classification setting. By the definition of variance of a multinomial distribution we can decompose the variance of $\widehat{\mathbf{p}}$ into 
\begin{equation}
\mathrm{Var}\big[\mathbf{\widehat{p}}\big] \approx \mathrm{tr}\Big(\mathbb{E}\big[\mathrm{diag}(\mathbf{\widehat{p}}) - \mathbf{\widehat{p}} \mathbf{\widehat{p}}^\top\big] + \mathbb{E}\big[\mathbf{\widehat{p}}^2\big] - \mathbb{E}\big[\mathbf{\widehat{p}}\big]^2\Big),
\end{equation}
where the first term represents aleatoric uncertainty $\sigma^2_{\mathrm{a}}$, and the second is the epistemic $\sigma^2_{\mathrm{e}}$. Each diagonal entry of the resulting matrix is the variance of a binomially distributed random variable, and the off-diagonals are negative covariances for fixed $T$. Since we're only interested in a single number to measure our uncertainty, we take trace of the resulting uncertainty matrix.

Alternatively, we can have the NN predict the input noise variance $\sigma^2_{\mathrm{a}}$ as one of its outputs~\cite{kendall2017uncertainties}, by assuming measurement error in our target function $y = \mathcal{F}(\mathbf{x}) + \epsilon$, where $\epsilon \sim \mathcal{N}(0,\sigma^2_{\mathrm{a}})$. The predictive variance in a multivariate classification setting is then given by
\begin{equation}
\label{uncertainty_}
\begin{split}
\mathrm{Var}\big[\bm{\widehat{y}}\big]  & \approx \frac{1}{T} \sum_t \exp({\widehat{\mathbf{s}}_t}) - \sum_j \mathbb{E}\big[\mathbf{\widehat{p}}\big] \log \mathbb{E}\big[\mathbf{\widehat{p}}\big]= \\ &= \mathbb{E}[\exp(\widehat{\mathbf{s}})] + \mathbb{H}\big[\mathrm{softmax}(\mathcal{F}(\mathbf{x};\bm{\widehat{\omega}}))],
\end{split}
\end{equation}
the entropy term measures epistemic uncertainty in the output softmax distributions, whereas the log aleatoric uncertainty $\widehat{\mathbf{s}}_i \coloneqq \log \sigma^2_{a,i}$ term is regressed by the NN for each input $\mathbf{x}_i$, for numerical stability. To capture aleatoric uncertainty in our classification task, we can use Monte Carlo integration on the NNs gaussian log-likelihood objective function, by drawing $t\in T$ samples of gaussian noise corrupted NN output logits $\mathcal{F}(\mathbf{x})$, yielding the following loss
\begin{equation}
\mathcal{L}_{\mathrm{NLL}} = -\log \mathbb{E}\big[\mathrm{softmax}(\mathcal{F}(\mathbf{x}) + \epsilon_t \odot \exp{(\widehat{\mathbf{s}}|\mathbf{x})})\big],
\end{equation}
with $\epsilon_t \sim \mathcal{N}(0,I)$ parameterised by the predicted aleatoric uncertainty $\exp{(\widehat{\mathbf{s}}|\mathbf{x})}$ for each sample $\mathbf{x}_i$, which learns to capture measurement error.

Having calculated the predictive uncertainty $\mathrm{Var}[\mathbf{\widehat{p}}]$ of our pseudo-labels, we calculate a per sample importance weight $\{\mathbf{x}_i, \widehat{\bm{y}}_i, \lambda_i\}$ with 
\begin{equation}
\label{eq: uncertainty_weight}
\lambda_i = \frac{1}{\exp(\mathrm{Var}[\mathbf{\widehat{p}}_i])^{\bm{\phi}(r)}},
\end{equation}
where $\bm{\phi}(\cdot)$ is a parameterised hyperbolic tangent function
\begin{equation}
\bm{\phi}(r) 
             = \frac{1 - \exp{(\gamma \cdot r + b)}}{1 + \exp{(\gamma \cdot r + b)}},
\end{equation}
with $\gamma$, $b$ as scale and intercept terms, and $r$ denotes the Self-Training iteration. The weighted penalised log likelihood of our NN with weights $\bm{\omega}$ is then
\begin{equation}
\label{PLL}
\mathcal{L}_{\mathrm{PLL}}= \sum_i \lambda_i \log p(\bm{y}_i|\mathbf{x}_i, \bm{\omega}) - \beta \mathbb{H}\big[p(\bm{y}_i|\mathbf{x}_i, \bm{\omega})\big],
\end{equation}
where $p(\bm{y}|\mathbf{x}, \bm{\omega})$ is computed via softmax, and the optional confidence entropy penalty term is balanced by $\beta$. By tuning $\gamma$ and $b$ we can obtain the desired behaviour over $r$ iterations, s.t. when the uncertainty is low we assign high weight to the predicted pseudo-labelled sample $\{\mathbf{x}_i, \widehat{\bm{y}}_i, \lambda_i \approx 1\}$. We can incrementally encourage the model to assign more weight to uncertain pseudo-labelled samples as Self-Training progresses, since in the $\lim_{r \to \infty} \phi(r)=-1$. Intuitively, this procedure inverts eq.~\eqref{eq: uncertainty_weight} over time, incrementally forcing exploration by adding more uncertain, and potentially informative samples, to the training set. In summary, using this logic along with entropy penalties on over-confident output distributions, we can mitigate the effect of pseudo-labelling error accumulation in the training set, and adjust risk taking by tuning $\gamma$ and $b$. 
\begin{figure*}[!t]
    \centering
    \includegraphics[width=\textwidth]{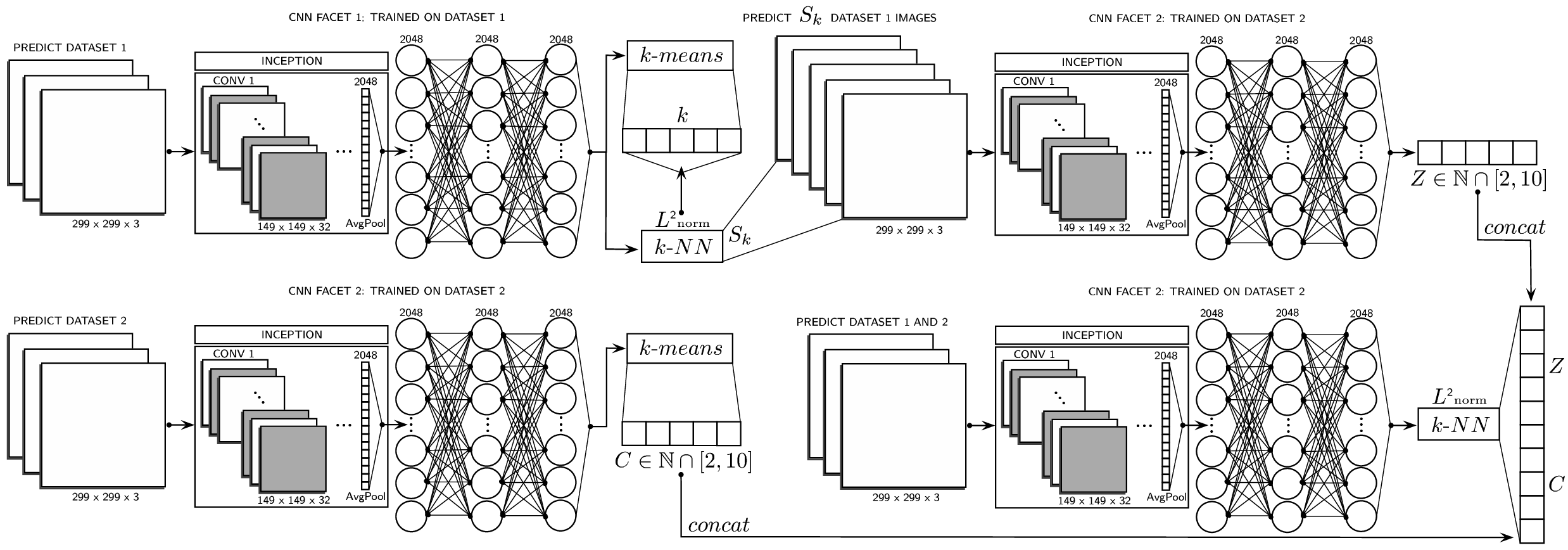}
\caption{Illustration of the multiple CNN facet adaptation framework proposed, which is based on clustering of extracted latent variable representations. The architectural details of each CNN are as previously described in Figure~\ref{fig: nn}.}
\label{fig: fc_kmeans}
\end{figure*}
Once per-sample predictive uncertainties are calculated, we decide on which pseudo-labelled samples to add to the training set via a Tukey Fence. Intuitively, assume a NN has been trained on data $\mathcal{D} = \big\{(\textbf{x}_i, \bm{y}_i)\}_{i=1}^N$, learning a function $\mathcal{F}(\mathbf{x};\bm{\omega})$ for mapping inputs to labels. At inference time, we take the correct predictions where $\bm{y}_i = \mathcal{F}(\mathbf{x}_i;\bm{\omega})$, and retrieve their predictive uncertainty. We then summarise variability by calculating the Interquartile Range (IQR) outlier statistic, and define an uncertainty upper bound $\tau$, under which pseudo-labelled samples from $\mathcal{U} = \big\{\widetilde{\textbf{x}}_i\}_{i=1}^{\widetilde{N}}$ have to be, in order to be added to $\mathcal{D}$ following
\begin{equation}
\mathcal{D}^* = \forall_i \in \big\{\mathcal{D} \cup \{\widetilde{\mathbf{x}}_i,  \widehat{\bm{y}}_i, \lambda_i\} \ | \ \mathrm{Var}[p(\bm{y}_i|\mathbf{x}_i)] < \tau\big\},
\end{equation}
where $\widehat{\bm{y}}_i$ denotes the pseudo-label assigned to sample $\mathbf{x}_i$ computed as $\widehat{\bm{y}}_i = \argmax \widehat{\mathbf{p}}_i$, and $\mathcal{D}^*$ is the augmented training set. Lastly, we can also easily adjust the uncertainty upper bound $\tau$ by selecting higher or lower quartiles to reflect how confident we'd like to be about predictions before adding samples to $\mathcal{D}^*$.
\section{\textbf{Latent Variable Adaptive Clustering}}
\label{subsec-Adaptation}
We propose a new simple methodology for visualising and analysing variability between distributions and attempt to adapt information from one problem to another in DNNs. In Figure~\ref{fig: fc_kmeans} an illustration of our adaptation framework is shown using an example backbone InceptionV3 CNN. Let the following denote $2$ training sets from separate datasets targeting the same task
\begin{equation}
\label{eq: training dataset}
\begin{split}
& \mathcal{D}_{1} = \big\{(\textbf{x}^{(i)}_{1}, \bm{y}^{(i)}_{1}) \ ; \ i=1,\ldots,N_1\big\}, \\
& \mathcal{D}_{2} = \big\{(\textbf{x}^{(i)}_{2}, \bm{y}^{(i)}_{2}) \ ; \ i=1,\ldots,N_2\big\},
\end{split}
\end{equation}
and the $2$ respective test sets as
\begin{equation}
\label{eq: test dataset}
\begin{split}
& \mathcal{T}_{1} = \big\{(\widetilde{\textbf{x}}^{(i)}_{1}, \widetilde{\bm{y}}^{(i)}_{1}) \ ; \ i=1,\ldots,\widetilde{N}_1\big\}, \\
& \mathcal{T}_{2} = \big\{(\widetilde{\textbf{x}}^{(i)}_{2}, \widetilde{\bm{y}}^{(i)}_{2}) \ ; \ i=1,\ldots,\widetilde{N}_2\big\}.
\end{split}
\end{equation}
Let $\mathcal{F}(\mathcal{D}_1; \mathbf{W}_1)$ and $\mathcal{F}(\mathcal{D}_2; \mathbf{W}_2)$ denote $2$ architecturally identical CNNs trained separately on each dataset. For each CNN we extract the final Fully-connected layer activations $\{\textbf{x}_{1}^{(i)}, \widetilde{\textbf{x}}_{1}^{(i)}\}\in\mathbb{R}^{2048}$ and $\{\textbf{x}_{2}^{(i)}, \widetilde{\textbf{x}}_{2}^{(i)}\}\in\mathbb{R}^{2048}$ as latent variables representations, by simply forward-propagating each image through as is typically done at inference time. 

Utilising these, our adaptation methodology is then performed as follows:
\begin{enumerate}
\item Given $\mathcal{D}_2$, produce a set of clusters $\mathbf{C} =\{\textbf{c}_1,\ldots,\textbf{c}_k\}$ by minimising the within-cluster $L^{2}$ norms of the following clustering objective function 
\begin{equation}
\label{eq:kmeans}
\widehat{\mathbf{C}}_{k\text{-means}} = \underset{\mathbf{C}}{\operatorname{arg\ min}} \sum_{i=1}^{k} \sum_{\mathbf{x}\in \mathbf{C}_{i}}^{} \big|\big|\mathbf{x}-\bm{\mu}_{i}\big|\big|^{2}.
\end{equation}
\item Repeat step 1 with $\mathcal{D}_{1}$ to generate $k$ clusters $\textbf{U} =\{\textbf{u}_1,\ldots,\textbf{u}_k\}$, and compute the $k$ closest instances in $\mathcal{D}_{1}$ to each centroid in $\textbf{U}$. Fetch the corresponding set of images $\textbf{S}=\{\textbf{S}_{1},\ldots,\textbf{S}_{k}\}$, whose latent variables are closest to $\textbf{U}$;
\item Forward-propagate $\textbf{S}$ through $\mathcal{F}(\mathcal{D}_2; \mathbf{W}_2)$ to obtain a new set of adapted clusters $\textbf{Z} =\{\textbf{z}_{1},\ldots,\textbf{z}_{k}\}$, where $\textbf{S}$ is considered an approximation of $\textbf{U}$ from $\mathcal{F}(\mathcal{D}_1; \mathbf{W}_1)$;
\item Derive an augmented cluster representation that encapsulates knowledge from both facets of the trained CNNs, by concatenating the respective $\textbf{C}$ and $\textbf{Z}$ clusters into a set $\textbf{A}=\{\textbf{c}_{1},\ldots,\textbf{c}_{k},\textbf{z}_{1},\ldots,\textbf{z}_{k}\}$;
\item Compute the euclidean distance between $\mathcal{T}_{1}$ and $\textbf{A}$ and evaluate the classification performance;
\item Iteratively remove the lowest performing cluster in $\textbf{A}$ and repeat step 5 until the performance stops improving.
\end{enumerate}
In all cases, the $k$-means++~\cite{arthur2007k} seeding strategy was used, whereby the first cluster center $\textbf{c}_1$ is chosen uniformly at random from $\mathcal{X}$, and all preceding cluster centers $\textbf{x}\in\mathcal{X}$ are chosen with probability
\begin{equation}
\textbf{c}_i = \frac{D(\textbf{x})^2}{\sum_{\textbf{x}\in\mathcal{X}}D(\textbf{x})^2},
\end{equation}
where $D(\textbf{x})$ denotes the distance between $\textbf{x}$ and the closest $\textbf{c}_i$. Moreover, we assign the class label of a given cluster $\textbf{c}_i$ as simply the mode class $j$ of all data points within it
\begin{equation}
\label{eq:cluster_label}
\textbf{c}_{i}^{j}=\max_{j\in J}{\big|\textbf{c}_{i} \cap j\big|}.
\end{equation}
In the experimental study of Section 6, we demonstrate that our method distills and adapts knowledge from both trained CNNs on real data, achieving better performance than direct inference of $\mathcal{T}_{1}$ with $\mathcal{F}(\mathcal{D}_2; \mathbf{W}_2)$, without any parameter retraining.
\section{Experimental Study}
\label{sec: experimental study}
This section is divided into two separate subsections, the first subsection presents experiments using Deep Bayesian Self-Training applied to the MNIST public domain dataset. An ablation study is presented and comparisons are made with baseline methods. The second subsection comprises a study using private (real) datasets, in which we perform some preliminary experiments using Transfer Learning and then we evaluate our proposed Latent Variable Adaptable Clustering method. We then finish off the second subsection by evaluating Deep Bayesian Self-Training on the self-annotation of the real datasets.
\subsection{\textbf{MNIST Dataset}}
In order to validate our algorithm we conduct a series of self-labelling experiments on the popular MNIST dataset. The MNIST dataset is comprised of 60,000 training, and 10,000 testing handwritten digit examples respectively. Firstly, we try to create a realistic scenario by splitting the 60,000 training examples into a smaller but balanced training set of only 50 examples per class, a validation set of 500 training examples per class, and allocate all remaining data to the unlabelled pool set. We begin by defining our backbone NN architecture of choice as a DenseNet~\cite{huang2017densely}. DenseNets have revealed several well founded advantages over previous architectures, from mitigating vanishing gradients, to encouraging feature propagation and reuse with shorter connections between layers~\cite{huang2017densely,jegou2017one}. The dense connectivity in DenseNets can be formally defined as
\begin{equation} \label{dense_connectivity}
\mathbf{A}^{[\ell]}= f\bigg(\text{BN}\Big(\mathbf{W}^{[\ell]} \cdot \big[\mathbf{A}^{[0]},\mathbf{A}^{[1]},\dots,\mathbf{A}^{[\ell-1]}\big]\Big)\bigg),
\end{equation}
where $f(\cdot)$ is the ReLU activation function, BN$(\cdot)$ is Batch Normalisation~\cite{ioffe2015batch} and $\big[\mathbf{A}^{[0]},\mathbf{A}^{[1]},\dots,\mathbf{A}^{[\ell-1]}\big]$ represents feature map-wise concatenation of all layers preceding $\ell$. A sequential composite function consisting of BN, ReLU and $3\times3$ convolution can then be defined as $H^{[\ell]}$. Each function $H^{[\ell]}$ produces $\omega$ feature maps, known as the growth rate of the network, and each layer $\ell$ takes as input $f + \omega\times(\ell-1)$ total feature maps, where $f$ denotes the number of channels in the visible layer. To reduce spatial dimensionality of feature maps, a Transition layer is introduced between densely connected DenseBlocks. Transition layers in~\cite{huang2017densely} are composed of BN followed by $1\times1$ convolution and $2\times2$ average pooling with a feature map compression factor $\theta = 0.5$. 

Following Algorithm 1 closely, we propose a progressively growing NN scheme by starting off with a 40 layer deep DenseNet with a growth rate $k = 12$, and incrementally increasing the growth rate (width) of the network as more data is added to the training set. In the first iteration, the network has only 181k parameters to avoid overfitting on the small initial training set, but complexity of the network is incrementally increased in an automated way. As described in greater detail in section~\ref{subsec: inverse uncertainty weighting}, we employ Monte Carlo Dropout at test time to calculate the predictive uncertainty of the assigned pseudo labels samples. In all cases, we take $T=30$ samples, equating to 30 different Dropout masks. We compare the performance of our proposed approach with a baseline ensemble method ($\mathrm{DEST}$) similar to~\cite{lakshminarayanan2017simple} for estimating predictive uncertainty, and the vanilla Self-Training methodology, albeit in a Deep Learning model, considering only the output probability of the NN as a measure of confidence, similarly to~\cite{lee2013pseudo}. We also evaluate the effect of our inverse uncertainty weighting scheme, as well as the entropy penalty on confident output distributions on the performance of our Bayesian Self-Training algorithm.
\begin{table*}[!ht]
    \label{table: dbst results}
    \centering
    \setlength{\extrarowheight}{2pt}
    \setlength{\arrayrulewidth}{.4pt}
    \caption{Deep Bayesian Self-Training results on self-labelling the MNIST dataset. $\tau$ is the upper bound uncertainty threshold for augmenting $\mathcal{D}^*$, $\lambda_i$ are sample-wise inverse uncertainty weights, and $r$ is the number of Self-Training iterations taken before stop conditions were met. All metrics (precision, recall, F1-score and cohen's $\kappa$) are reported in $1-$metric format.}
    \begin{tabular}{c|c|c|c|c|c|c|c|c|c}
    \hline \hline
    \multicolumn{10}{c}{Deep Bayesian Self-Training ($\mathrm{DBST}$) Results on MNIST} \\
     \hline
    Model & $\tau$ & $\lambda_i$ & $\mathcal{L}_{\mathrm{PNLL}}$ & Precision & Recall & F1-score & Unlabelled & Cohen's $\kappa$ & $r$ iters \\
    \hline
    $\mathrm{DST}$ & - & \xmark & \xmark & .0103 & .0103 & .0103 & 781 & $\mathbf{.0115}$ & 15 \\
    $\mathrm{DEST}$ & Q3 & \cmark & \xmark & .0044 & .0044 & .0044 & 4,391 & .0049 & 20 \\
    \hline
    $\mathrm{DBST}$-1 & Q3 & \xmark & \xmark & .0042 & .0043 & .0043 & 5,044 & .0045 & 15 \\
    $\mathrm{DBST}$-2 & Q3 & \cmark & \xmark & .0032 & .0032 & .0032 & 4,092 & .0035 & 21 \\
    $\mathrm{DBST}$-3 & Q2 & \cmark & \xmark & .001 & .001 & .001 & 17,828 & $\mathbf{.0011}$ & 27 \\
    $\mathrm{DBST}$-4 & Q2 & \cmark & \cmark & .0071 & .007 & .0071 & \textbf{762} & .0079 & 26 \\
    \hline \hline
    \end{tabular}
    \label{results table}
\end{table*}
\subsubsection{Training Details}
In all MNIST experiments, we use the same DenseNet model and hyperparameters for fair comparisons. Specifically, we train the networks using Stochastic Gradient Descent (SGD) with a Nesterov momentum of $0.9$, a batch size of 32, and an initial learning rate of $0.1$. We train all models for $75$ epochs and reduce the learning rate by a factor of 10 at 50 and 75\% of the way through training. All models are trained using the same train/valid/test/unlabelled splits and no data augmentation is used aside from simple image standardisation (mean 0 sd. 1) and we take $T=30$ Monte Carlo Dropout samples to at test time as explained in section~\ref{subsec: inverse uncertainty weighting}. With regards to the ensemble, we train $M=5$ models each initialised with random weights, and capture the predictive uncertainty following equation~\eqref{uncertainty_}, but without using Dropout at test time. Lastly, the stop conditions can be adjusted depending on the application at hand, but here they were kept consistent in all experiments for fairness of comparison. Specifically, we stipulate that if less than the current batch size number of images are selected to be added to the training set in the next Self-Training iteration, the algorithm stops. 
\begin{figure}[b]
    \centering
    \setlength{\tabcolsep}{0.1em}
    \includegraphics[trim={0 0 0 0},clip,width=0.48\textwidth]{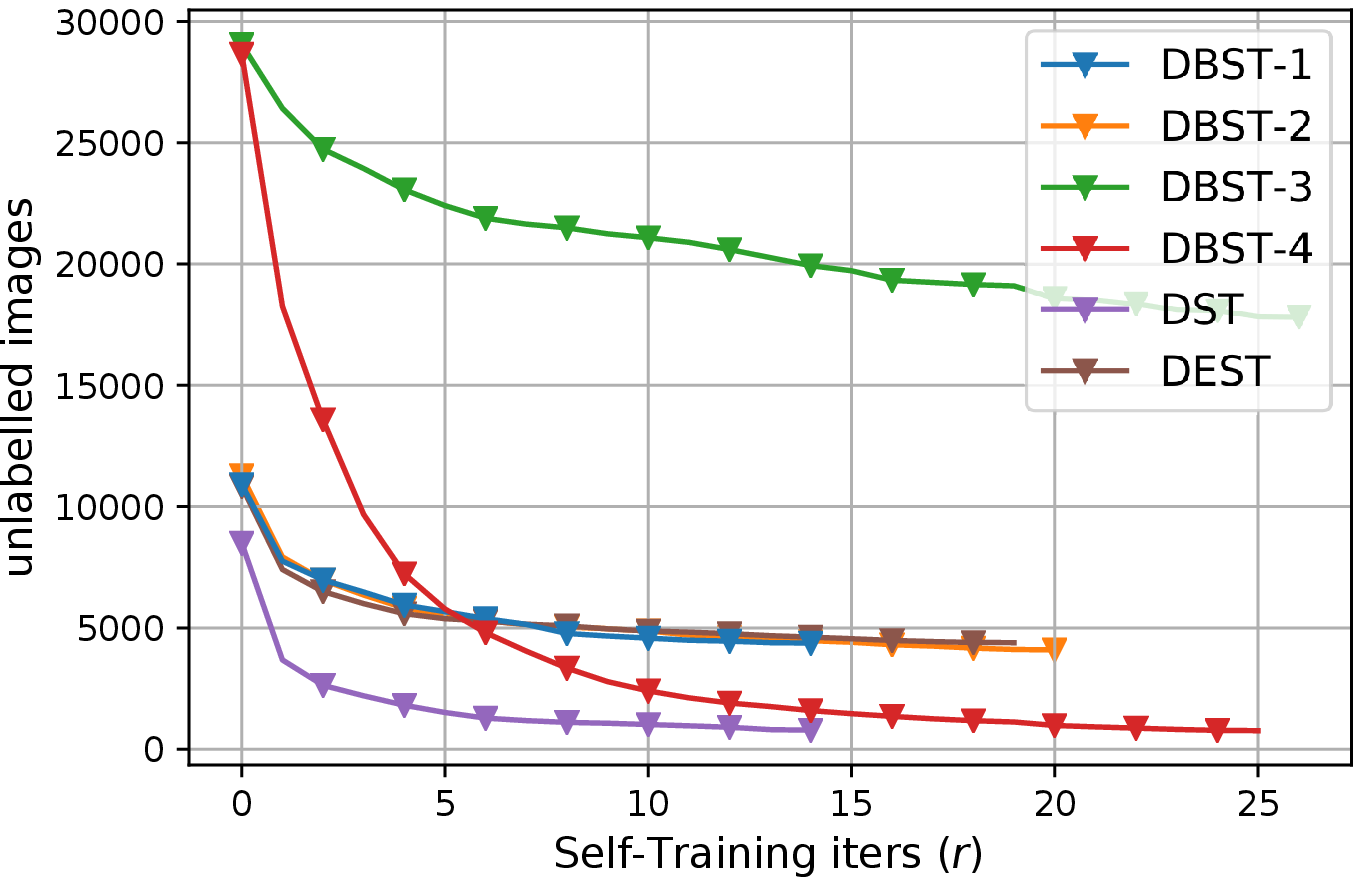}
 \caption{Self-Training model comparisons regarding number of images left unlabelled after $r$ iterations. Notice how the baseline Self-Training (DST) is overconfident by wrongly pseudo-labelling more samples early and propagating these errors, resulting in a lower cohen's $\kappa$ score as reported in Table~\ref{results table}.}
\label{samples}
\end{figure}
\begin{figure*}[!t]
    \centering
    \setlength{\tabcolsep}{0.1em}
    \begin{tabular}{cc}
        \subfloat[]{\includegraphics[trim={0 0 0 0},clip,width=0.5\textwidth]{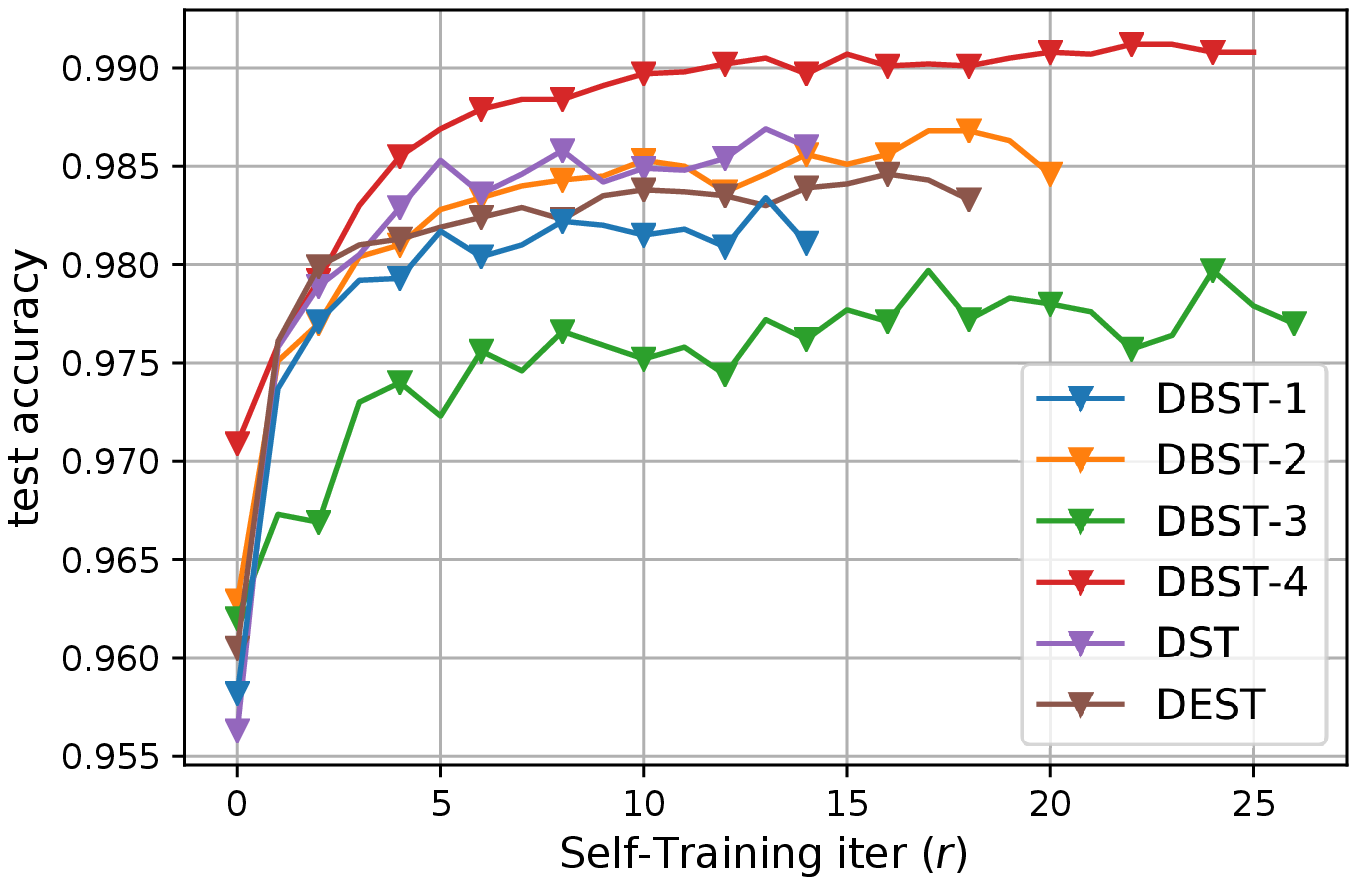}\label{test acc}}&
        \subfloat[]{\includegraphics[trim={0 0 0 0},clip,width=0.5\textwidth]{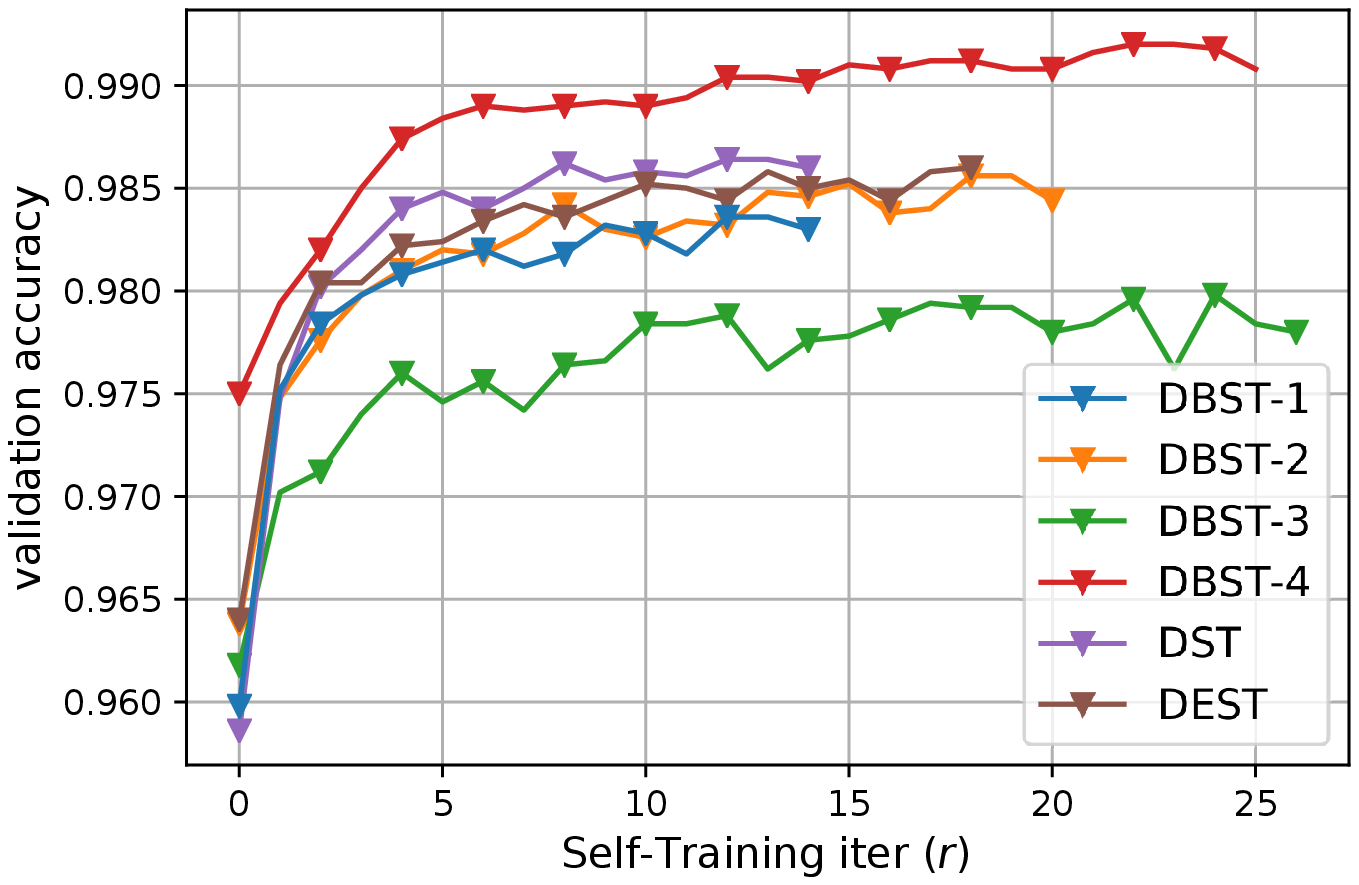}\label{valid acc}}
    \end{tabular}
 \caption{Model performance comparisons over $r$ Self-Training iterations. \textbf{(a)} MNIST test set performance after each Self-Training iteration. \textbf{(b)} As in (b) but comparing validation set performance. Notice that every model uses the same stop condition for fair comparison, but they stop at different times due to their uncertainty level. DBST-4 using both inverse uncertainty sample-weights and an entropy penalty on the log-likelihood loss ($\mathcal{L}_{\mathrm{PNLL}}$), generalises better as reported Table~\ref{results table}.}
\label{accs}
\end{figure*}
\begin{figure*}[t]
    \centering
    \setlength{\tabcolsep}{0.1em}
    \begin{tabular}{cc}
        \subfloat[]{\includegraphics[trim={0 0 0 0},clip,width=0.5\textwidth]{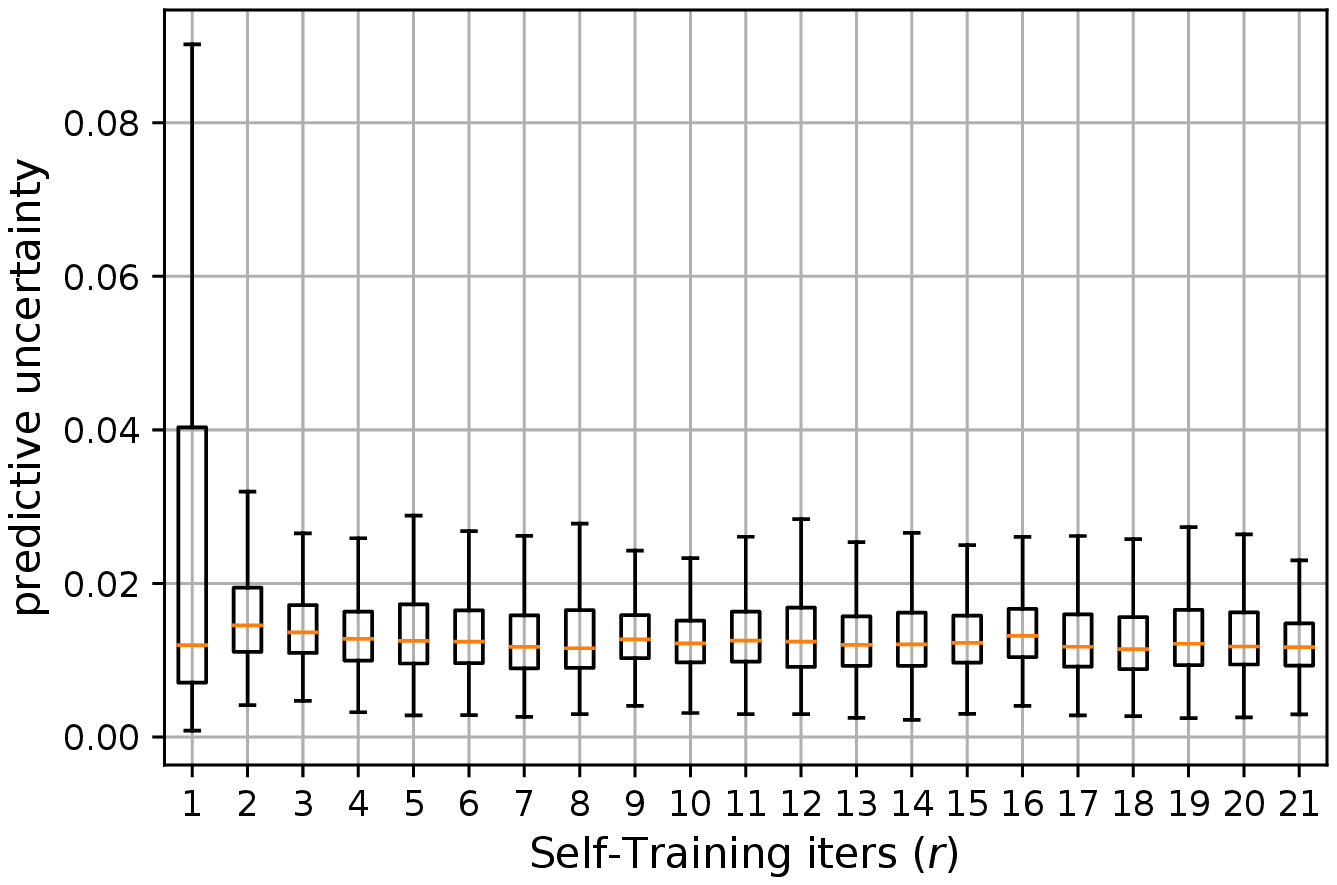}\label{1}}&
        \subfloat[]{\includegraphics[trim={0 0 0 0},clip,width=0.5\textwidth]{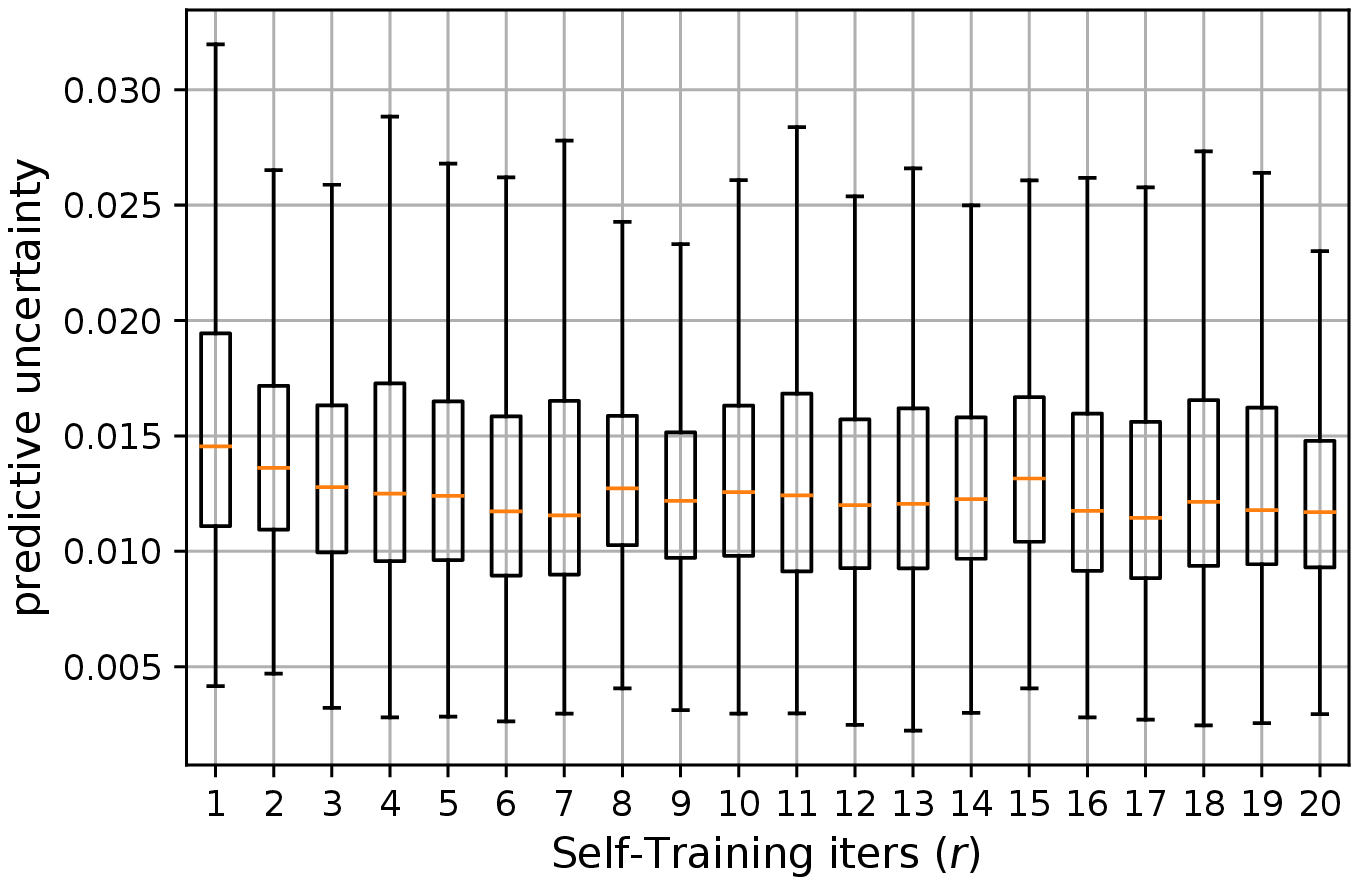}\label{2}}
    \end{tabular}
 \caption{Box plots (IQR) depicting the quartiles for setting the uncertainty upper bound threshold threshold $\tau$ over $r$ iterations in the $\mathrm{DBST}$-2 model as an example. Note: these IQR stats are calculated using the predictive uncertainies of correctly classified samples in the train/valid/test sets only. \textbf{(a)} Shows all iterations ($r=21$) whereas \textbf{(b)} omits the first one for better visibility.}
\label{iqr}
\end{figure*}
\begin{figure*}[ht]
    \centering
    \begin{tabular}{ccccc}
        \subfloat[]{\includegraphics[trim={98 10 98 0},clip,width=0.19\textwidth]{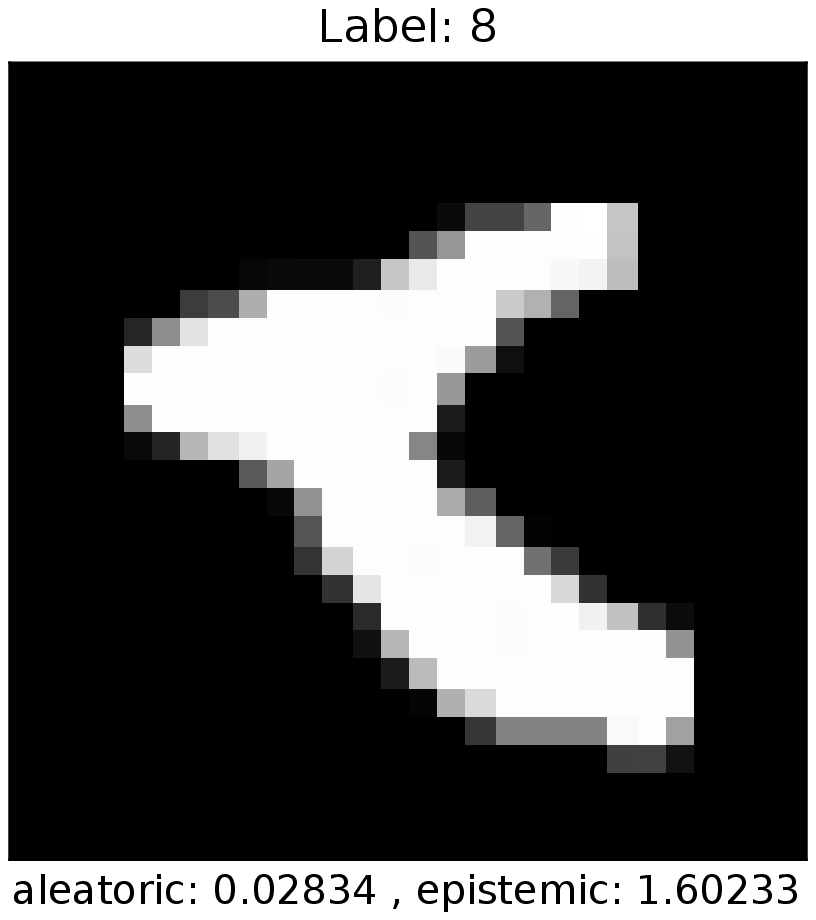}}&\hspace{-1.2em}
        \subfloat[]{\includegraphics[trim={98 10 98 0},clip,width=0.19\textwidth]{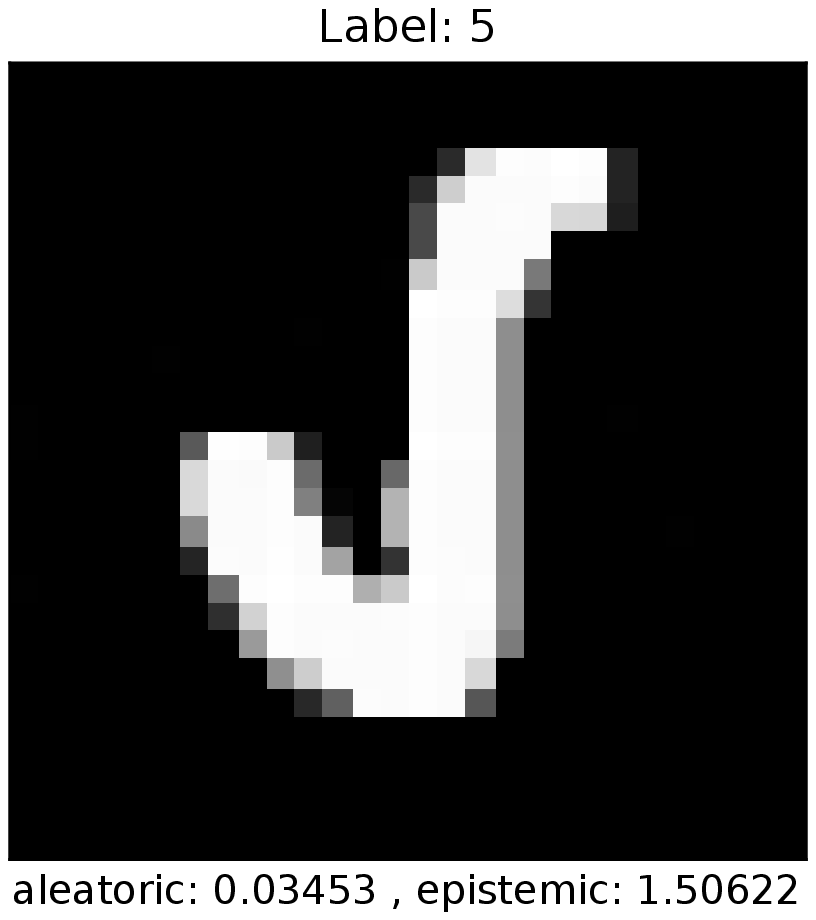}}&\hspace{-1.2em}
        \subfloat[]{\includegraphics[trim={98 10 98 0},clip,width=0.19\textwidth]{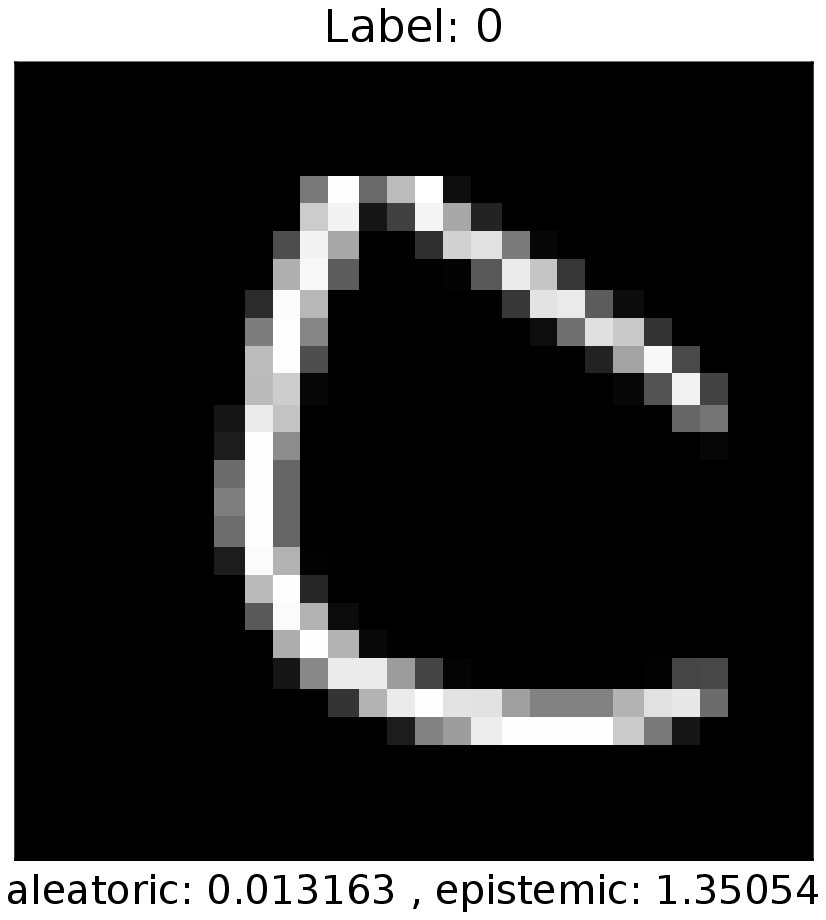}}&\hspace{-1.2em}
        \subfloat[]{\includegraphics[trim={98 10 98 0},clip,width=0.19\textwidth]{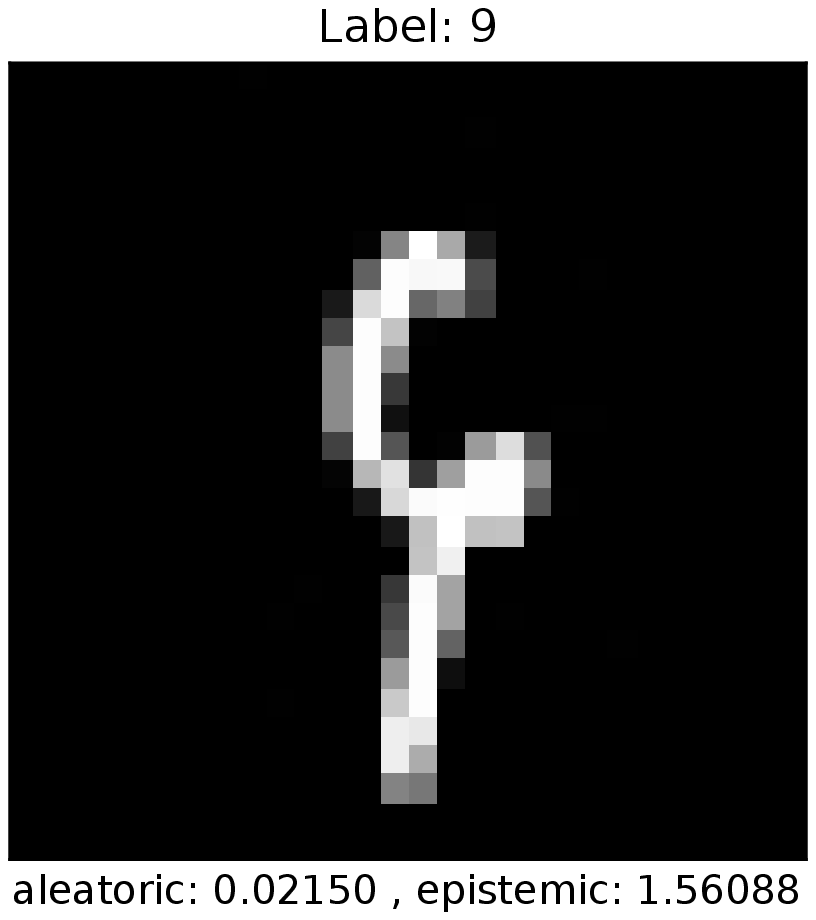}}&\hspace{-1.2em} 
        \subfloat[]{\includegraphics[trim={98 10 98 0},clip,width=0.19\textwidth]{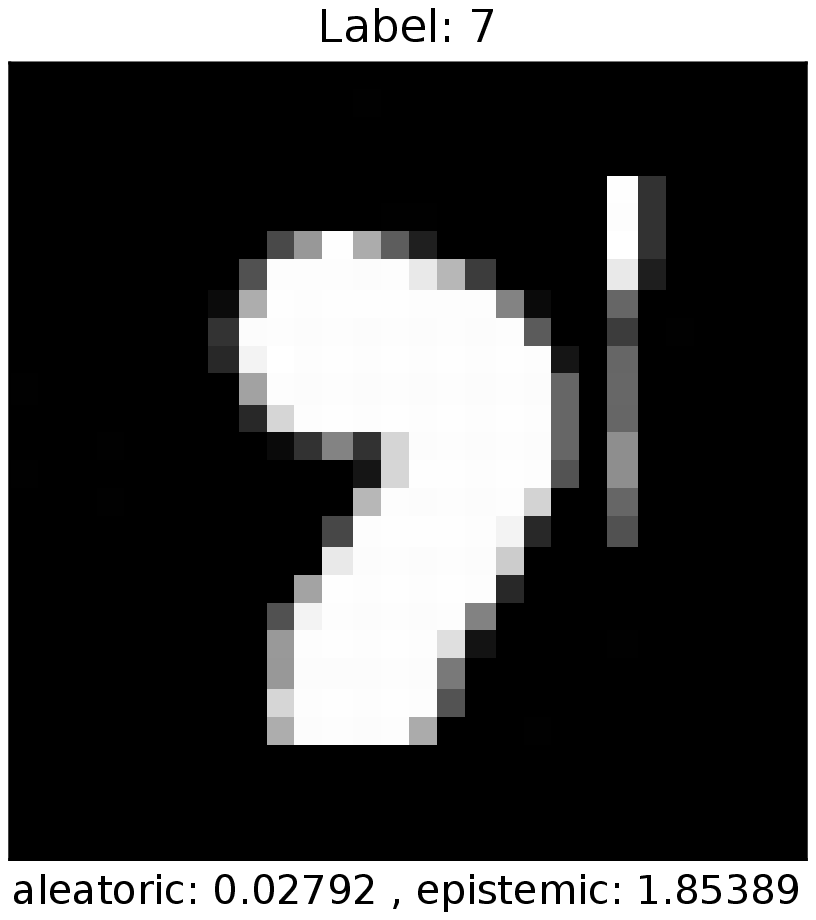}}
    \end{tabular}
\caption{Examples of images left in the unlabelled pool set for model $\mathrm{DBST}$-2. Images with the highest epistemic uncertainty were selected for each digit class, along with their corresponding aleatoric uncertainties reported in the x-axis. The actual label of each image is found on top. As we can see from these difficult examples, these digits were automatically identified as problematic (too uncertain) in the $\mathrm{DBST}$ pseudo-labelling process, so they were not added to the training set $\mathcal{D}^*$.}
\label{images_unlabelled}
\end{figure*}
\subsubsection{Ablation Study}
The results are reported in Table~\ref{results table}, and illustrated in Figures~\ref{images_unlabelled}, ~\ref{accs}, \ref{iqr} and \ref{samples}. In our experiments we simply have the NNs predict the labels for the 54,500 unlabelled MNIST samples, and evaluate how well the system is doing at predicting the correct labels at the end of each Self-Training iteration. The evaluation is primarily considered in terms of the cohen's kappa statistic ($\kappa$) as it is more robust than accuracy by taking into account random luck, and the number of images left unlabelled after Self-Training. As can be observed from the results, the addition of our proposed inverse uncertainty weighting scheme improves the performance of the algorithm by leaving less images unlabelled, and achieving a higher $\kappa$ score (DBST-1 to DBST-2). 

We also test the effect of the quartile uncertainty thresholds for $\tau$ from Q3 to Q2 (DBST-2 to DBST-3), meaning we are more strict about which pseudo-labelled samples we can add to the training set. This only considers very highly condident pseudo-label predictions resulting in a higher $\kappa$ score, at the cost of labelling less examples as expected. 

In the DBST-4 model, we combine both the sample-wise inverse uncertainty weighting scheme and the entropy penalty on the log likelihood loss ($\mathcal{L}_{\mathrm{PNLL}}$) using $\beta=1$ as described in section~\ref{subsec: inverse uncertainty weighting}. As reported in Table~\ref{results table}, the number of examples left unlabelled is significantly less, whilst maintaining a good cohen's $\kappa$ agreement between predicted and actual labels. In comparison to the others, the DBST-4 model provides the best balance between the number of unlabelled images left after self-training and a high cohen's $\kappa$ score.
\subsubsection{Comparitive Discussion}
Lastly, we compare our Bayesian models ($\mathrm{DBST}$) with two baseline method for estimating uncertainty in a similar way to~\cite{lakshminarayanan2017simple}, known as a Deep Ensemble of NNs ($\mathrm{DEST}$), and the standard Self-Training ($\mathrm{DST}$) following the logic in~\cite{lee2013pseudo}, and simply using the NNs predicted probability of an assigned pseudo-label as a level of confidence. The predictions from each NN in the ensemble ($\mathrm{DEST}$) can be used as to calculate predictive uncertainty as the deviations capture model parameter uncertainty. Here we do not employ any bootstrap methods as the randomness from the NN weight initialisation and shuffled training has been shown to be sufficient experimentally~\cite{lakshminarayanan2017simple}. We use the same DenseNet architecture, including related hyperparameters and identical dataset splits to train an ensemble of 5 models. Table~\ref{results table} shows that our methods (DBST) are better than using an ensemble ($M=5$) for predicting uncertainty for our Self-Training purpose, whilst taking approximately 5$\times$ less time to run in our experiments. Note that Monte Carlo Dropout samples are very cheap to compute at inference time compared to training multiple models, thus we can afford to take multiple samples i.e. $T=30$ as compared to an ensemble of $M=5$, which is also an advantage of our approach. 

With regards to the vanilla Self-Training baseline ($\mathrm{DST}$), again we use the exact same DenseNet architecture and related hyperparameters for fair comparisons. As previously outlined, in standard Self-Training we take the NNs predicted probability as a measure of confidence, and to demonstrate the inadequacy of this method we threshold with a very high confidence probability of $.99$. This simply means that only pseudo-label predictions above the $.99$ probability (confidence) threshold in a 10-way softmax (MNIST digit classes) are added to the training set. As reported in Table~\ref{results table} and Figure~\ref{samples}, $\mathrm{DST}$ under performs compared to our methods since it is over confident early on, resulting in the addition of more wrong pseudo-labels to the training set thus propagating the errors forward. Although the number of images left unlabelled is low, the cohen's $\kappa$ score is significantly lower 
\begin{figure*}[!t]
    \subfloat{
        \includegraphics[width=.99\columnwidth]{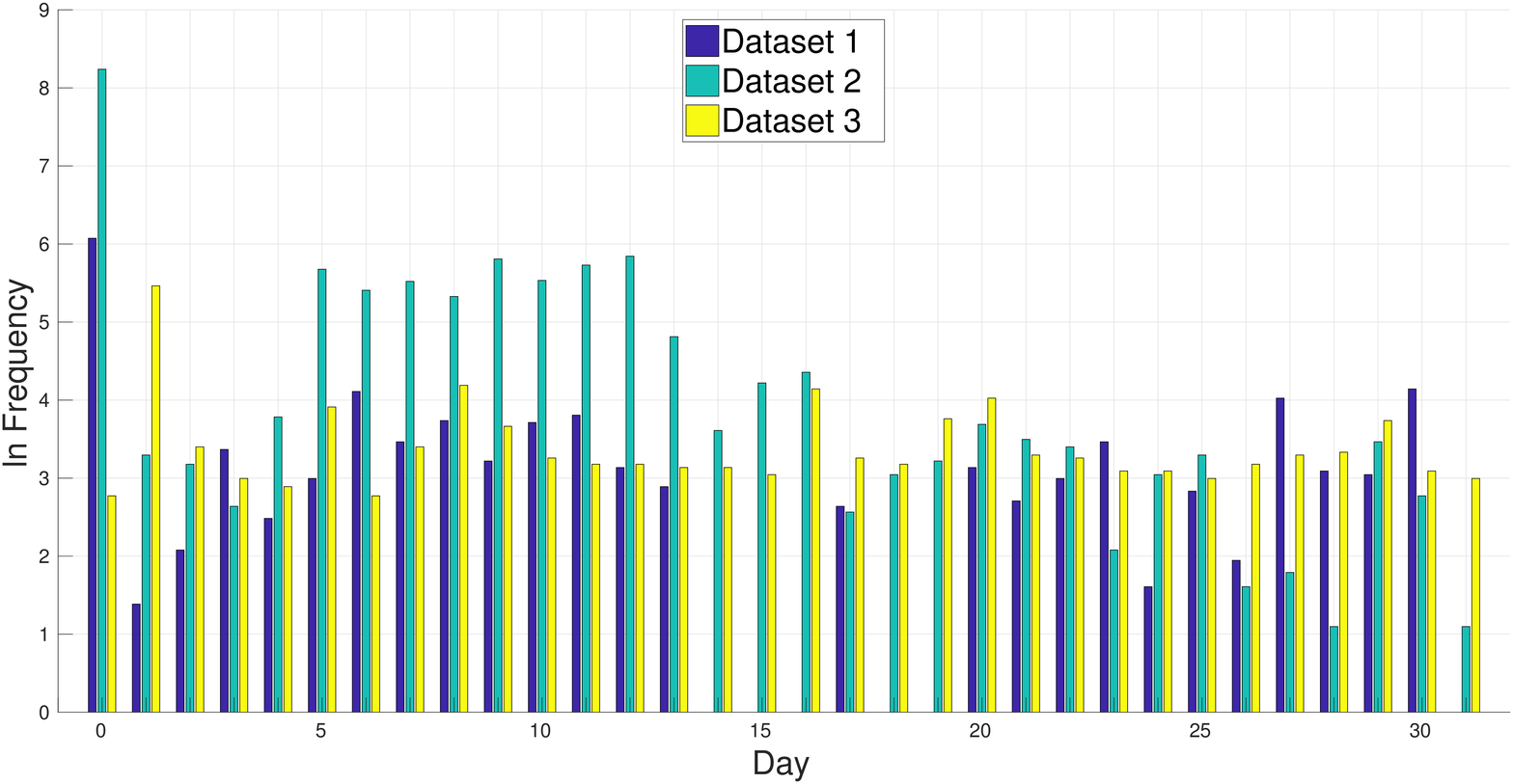}} \quad \
    \subfloat{
        \includegraphics[width=.99\columnwidth]{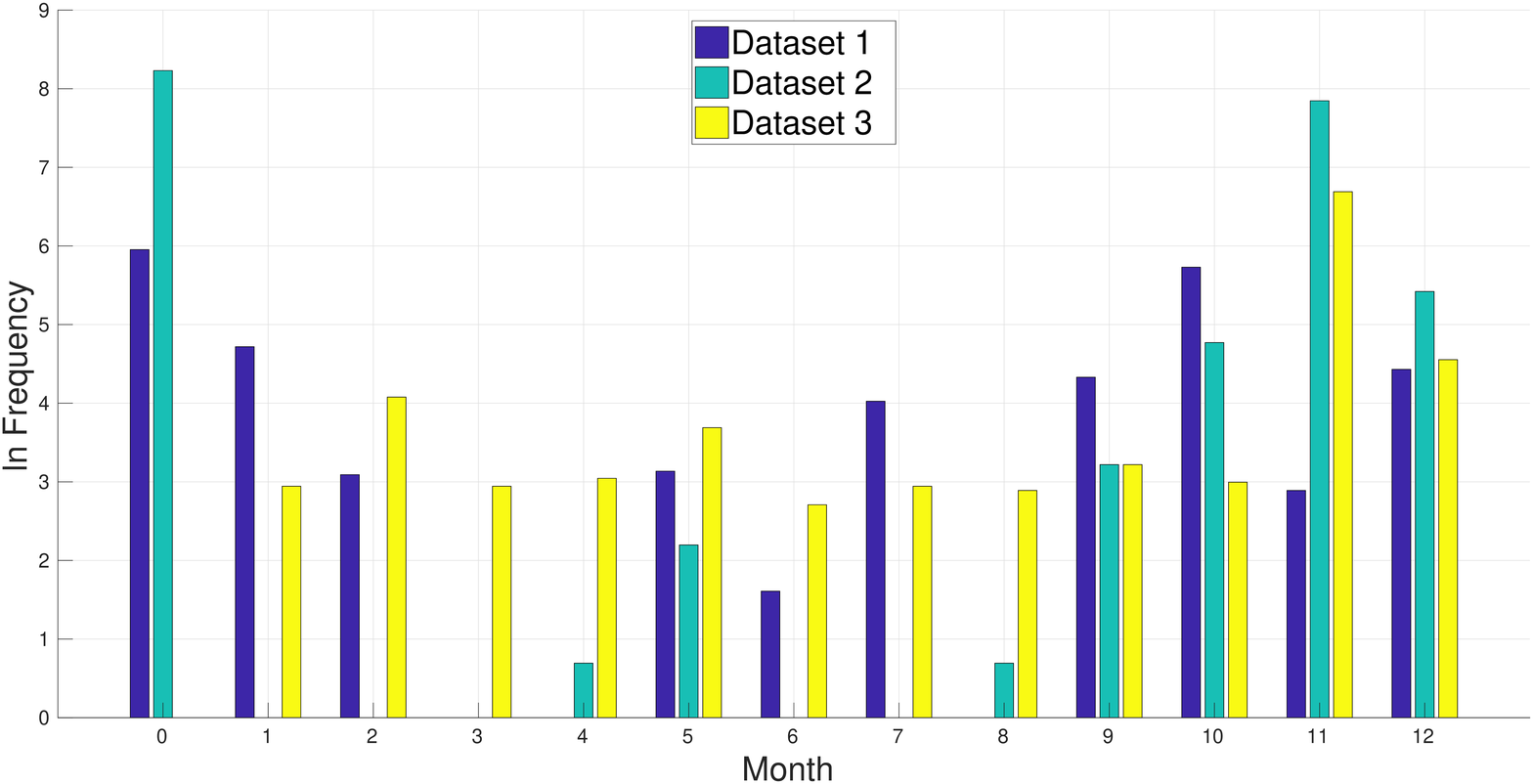}}
    \caption{\textbf{Left:} Frequency ($\ln$ scale) of appearance per `Day' in \textit{use-by} dates. \textbf{Right:} Respective appearance per `Month'.}
    \label{fig: histogram}
\end{figure*}
\subsection{\textbf{Real Datasets}}
\begin{figure*}[t]
    \centering
    \begin{tabular}{ccccc}
        \subfloat[]{\includegraphics[width=0.177\textwidth]{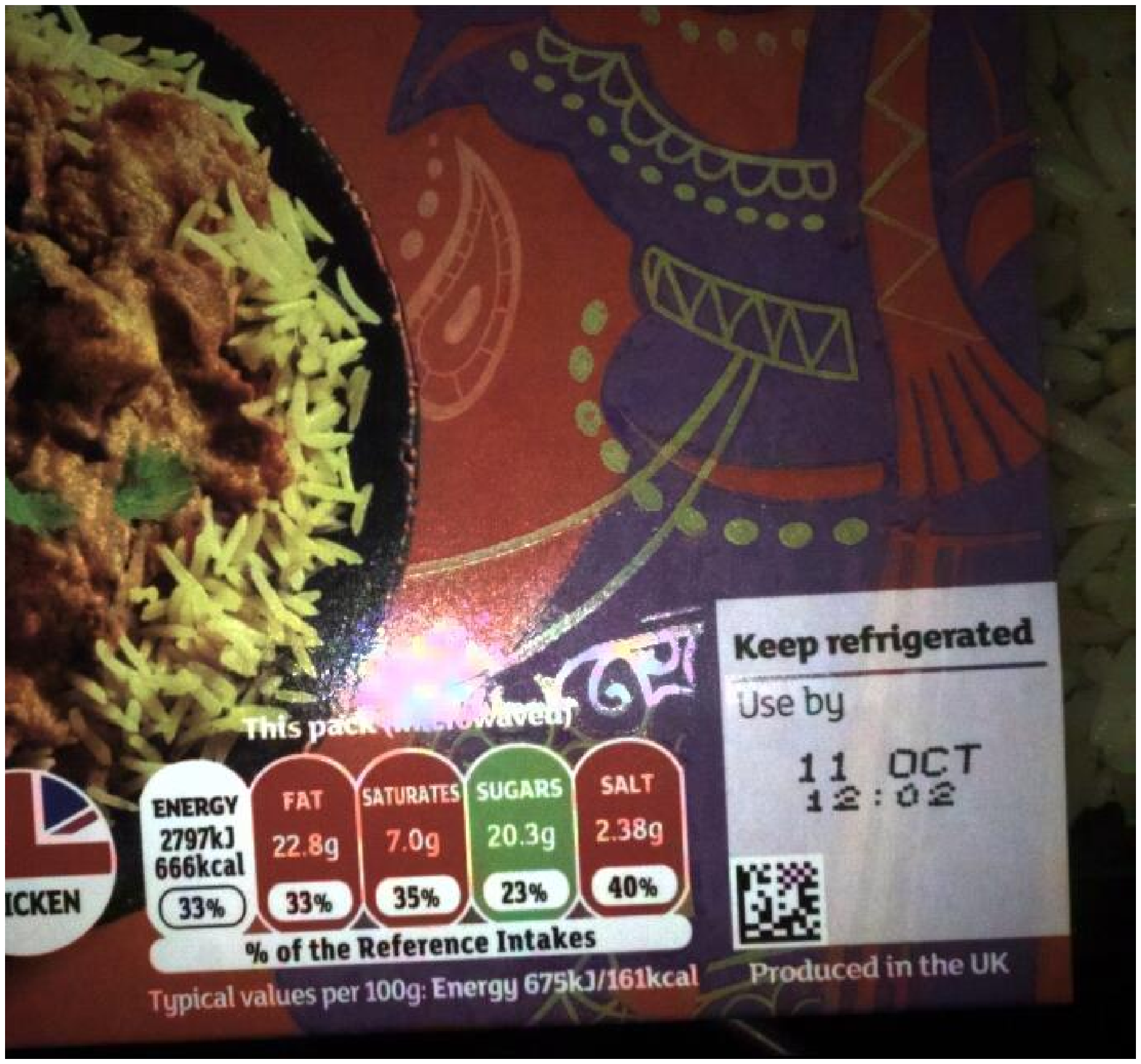}}&
        \subfloat[]{\includegraphics[width=0.177\textwidth]{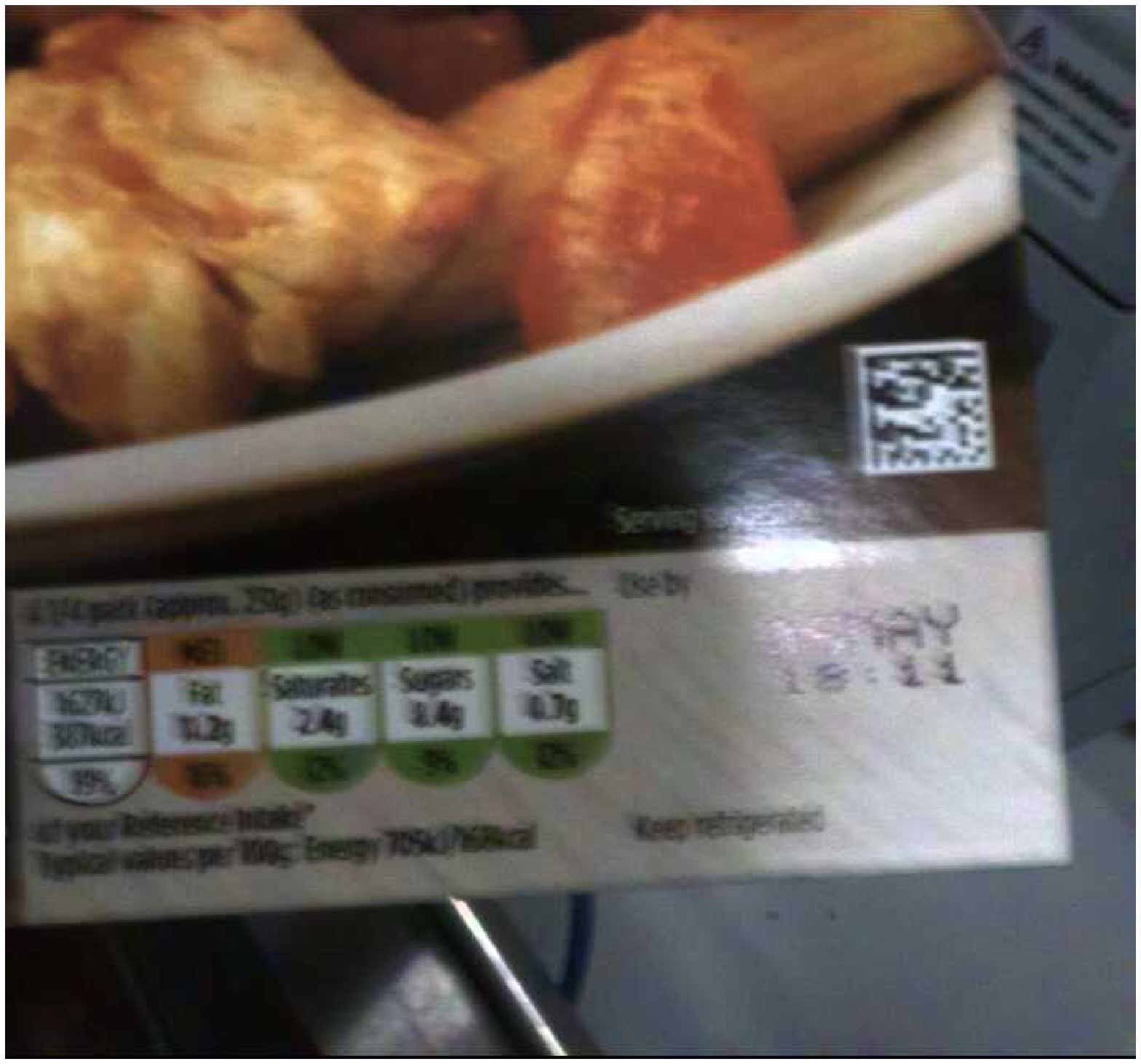}}& 
        \subfloat[]{\includegraphics[width=0.177\textwidth]{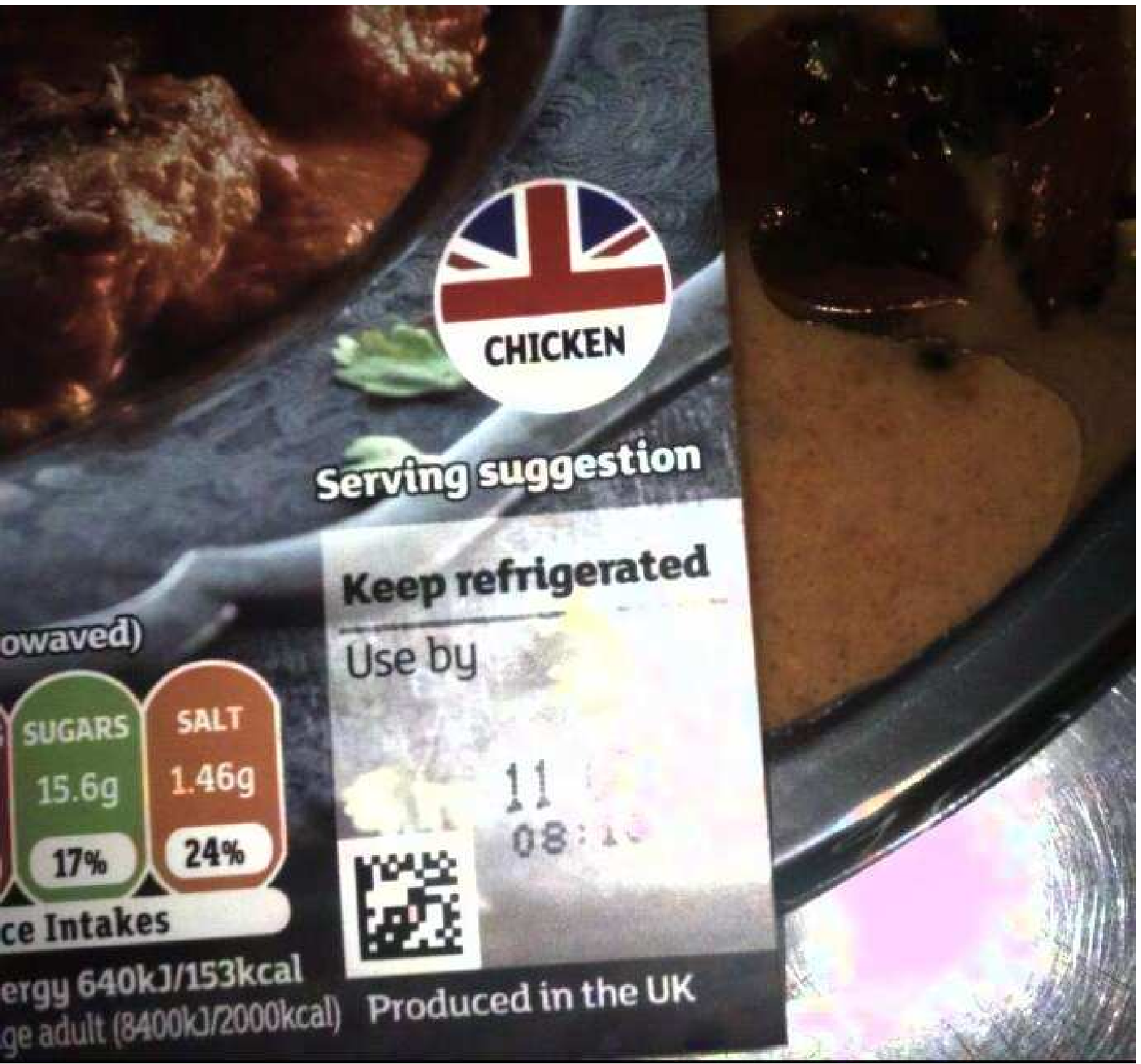}}& 
        \subfloat[]{\includegraphics[width=0.177\textwidth]{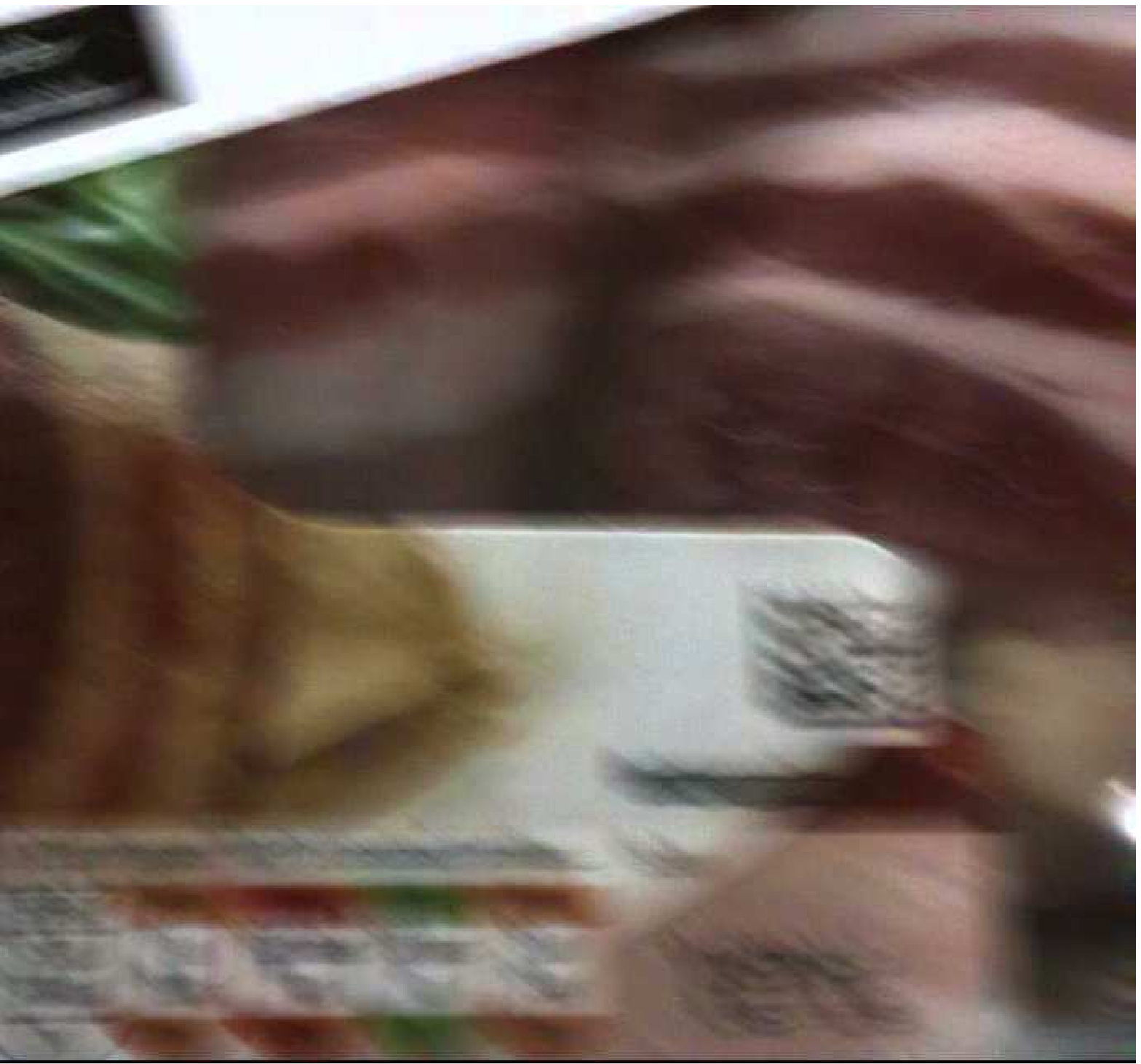}}& 
        \subfloat[]{\includegraphics[trim={0 0 115 0},clip,width=0.177\textwidth]{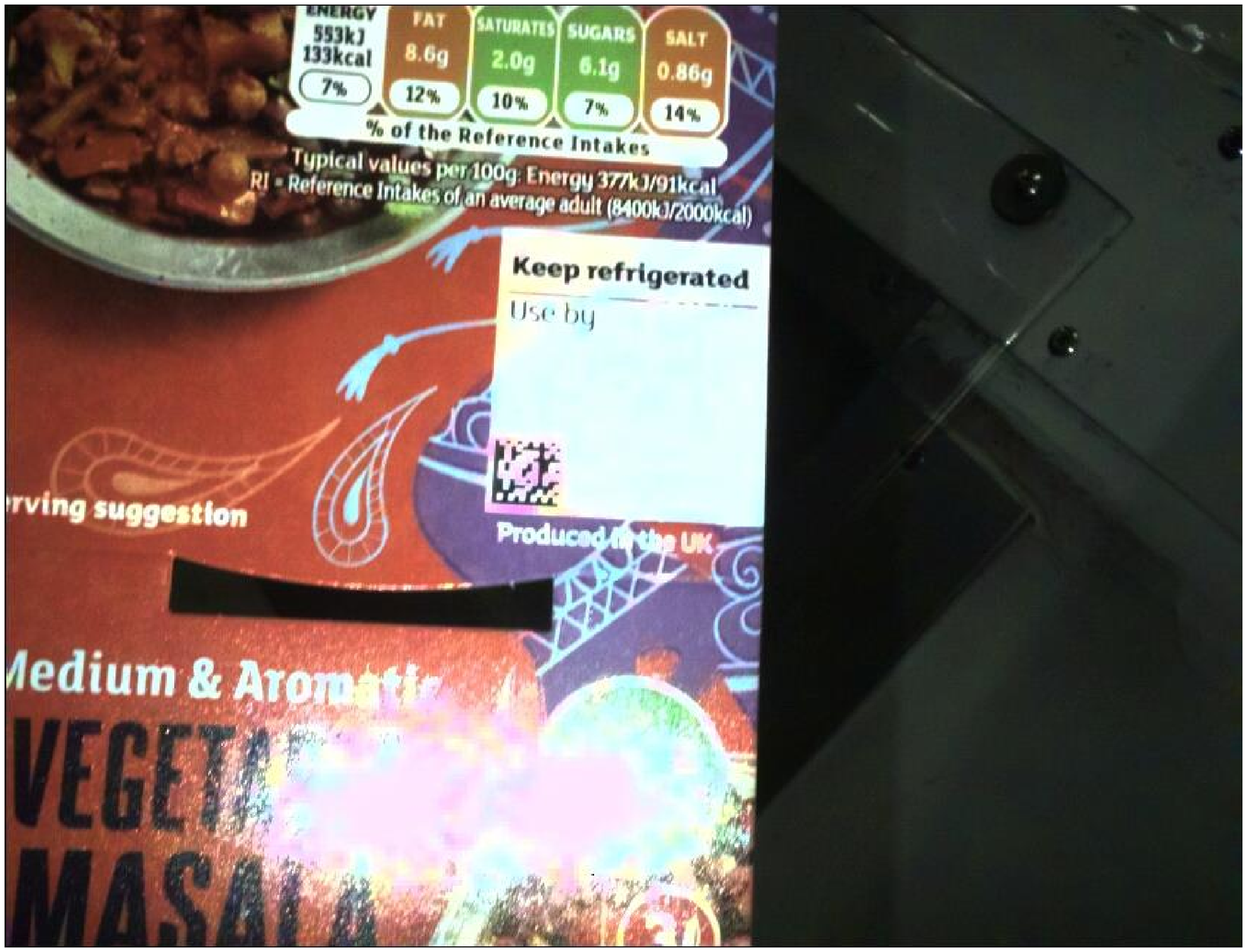}}
    \end{tabular}
\caption{Per category examples of images in our datasets. \textbf{(a)} Complete Date (day and month visible). \textbf{(b)} Partial Date (no day visible). \textbf{(c)} Partial Date (no month visible). \textbf{(d)} Unreadable. \textbf{(e)} No date (neither day or month visible).}
\label{table: table chategories}
\end{figure*}
\label{sec: dataset pre-processing}
Four datasets of food package photographs were collected by a leading food company and provided to us for research purposes. The four sets include $1404$, $6739$, $1154$ and $13948$ captured images respectively. In order to produce trainable datasets, a portion of the images was first manually annotated w.r.t. the presence of \textit{use-by} dates, and lack thereof. In the case of unreadable images, in which dates were not discernible from the the background - potentially due to heavy distortion, non-homogeneous illumination or blur - were then set aside in a separate category. Conversely, images in which either day or month, or both were missing, were considered as incomplete, and subsequently grouped into their own category. Lastly, images of good quality, reporting the date including both the day and month, were considered as good candidates for OCV. The first three sets of images were annotated as mentioned above to form 5 categories: complete dates, missing day, missing month, no date and unreadable (Table~\ref{table:dataset details}), whereas photographs belonging to the fourth dataset were annotated as good or bad candidates for OCV, and utilised to test our proposed Bayesian self-annotating framework. After annotating all the images in the first three datasets, it was possible to plot some statistics (see Figure~\ref{fig: histogram}) on the frequency of specific dates within each dataset, and thus devise a methodology for conducting experiments with balanced sets of classes. Moreover, by inspecting the images with partially missing data, it was observed that most of them were photographs of package labels which had been folded at crucial points, included photographic glare, digits fainting over time, or included human made occlusions. With regard to the fourth dataset, 8931 images were annotated as including readable dates, and the remaining 5017 as unreadable.
\subsubsection{Transfer Learning}
\begin{figure}[!t]
    \includegraphics[width=\columnwidth]{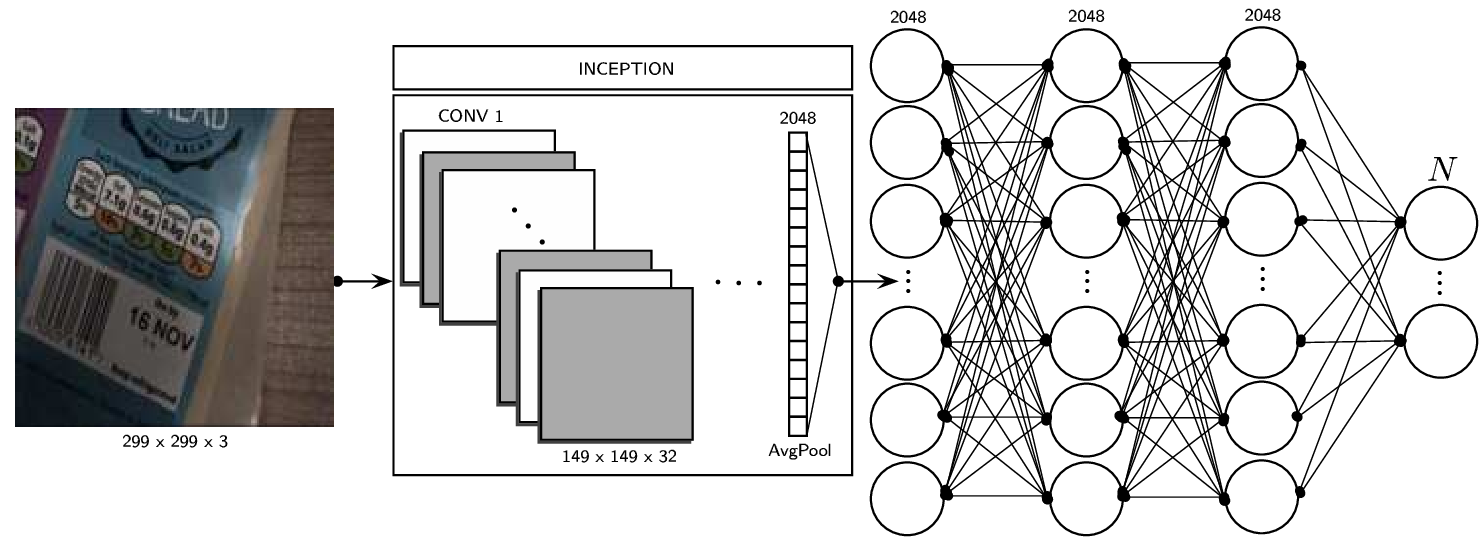}
\caption{Depiction of the classification architecture. From left to right, input images were resized to $299 \times 299 \times 3$ to accommodate the CNN's convolutional layer parameters and arithmetic. There exist 2 hidden layers with 2048 units each and ReLu activations. The number of units $N$ in the softmax layer was adjusted as per the number of classes being classified in different experiments.}
\label{fig: nn}
\end{figure}

It was of particular interest to conduct transfer learning in order to assess the adaptability of pre-trained CNN weights~\cite{szegedy2016rethinking} on the current food datasets. Specifically, each image from our datasets was fed through a previously trained InceptionV3 CNN on the ImageNet dataset, up to the last global average pooling (GAP) layer, where a $2048$ dimensional vector representation of each instance was extracted. The $2048$ dimensional vectors then became the input to a new series of FC layers and a final softmax layer able to predict $N$ classes (see Figure~\ref{fig: nn}). In order to optimise the training performance of the new FC layer network, a series of architectural decisions were made empirically, and the best performances were achieved using a FC network consisting of two $2048$ unit hidden layers with Rectified Linear Unit (ReLU) activations and Batch Normalization (BN)~\cite{ioffe2015batch} layers. 

The risk of overfitting rises as the number of parameters increases w.r.t. number of training examples. Due to the limited amount of training data, available for experimentation, it is infeasible to train state-of-the-art models from scratch. Therefore, we introduced an effective regulariser in the new network as well as adapted previously learned low-level features through transfer learning. One of the most effective regularisation techniques is Dropout~\cite{srivastava2014dropout}. In practice, to preserve more information in the input layer $\ell^{(0)}$ (of $L$ total layers) in the network and thus aid learning, the probability of keeping ($p(z^{(i)}):\neq0$) any given neuron $z^{(i)}$ in layer $i$ was as defined per the following schema
\begin{equation}
    \ell^{(i)}=\begin{cases}
               p(z^{(i)})=0.8 & \text{if~~$i=0$}\\
               p(z^{(i)})=0.5 & \text{otherwise}.
\end{cases}
\end{equation}
In view of the unbalance present among the various classes, it was beneficial to use a weighted negative log-likelihood as a loss function \eqref{eq:  crossentropy}. In \eqref{eq:  crossentropy}, $\lambda_{j}$ is a weight coefficient computed for the $j^{th}$ of all classes $J$ as a function of the proportion of instances $N_{j}$ compared to the most densely populated class \eqref{equation: weight}. During training, $\lambda$ encourages the model to focus on under-represented classes
\begin{equation}
  \label{eq:  crossentropy}
  \mathcal{L}_{\mathrm{NLL}} = - \sum_{i}\lambda_{j}\bm{y}_i\log ({\widehat{\bm{y}}_i}) - (1-\bm{y}_i)\log(1-{\widehat{\bm{y}}_i})
\end{equation}
calculating the per-class weight parameter $\lambda_j$ with
\begin{equation}
  \label{equation: weight}
  \lambda_{j} = \frac{1}{N_{j}}\max\Big(\big\{N_{i}\big\}_{i=[1:J]}\Big).
\end{equation}
In the case of multiclass classification, where $J > 2$, the weighted cross entropy loss function can be defined as
\begin{equation}
  \label{eq:  categorical_crossentropy}
   \mathcal{L}_{\mathrm{NLL}} = -\sum_{i=1}^{M}\sum_{j=1}^{J}{\lambda_j\bm{y}_{ij}\log ({\widehat{\bm{y}}_{ij}}}),
\end{equation}
where $\log p(\widehat{\bm{y}} = j | \bm{z}_j)$ is calculated as
\begin{equation}
\log \mathrm{softmax}(\bm{z}_j) = \log \Bigg[\frac{\exp(\bm{z}_j)}{\sum_{k}\exp(\bm{z}_k)}\Bigg],
\end{equation}
$\bm{z}$ is a vector of NN output logits, and M denotes the batch size of choice for stochastic optimisation of $\mathcal{L}_{\mathrm{NLL}}$ via backpropagation. In all cases, we use Adaptive Moment Estimate (Adam) as an optimiser~\cite{kingma2014adam}. 

In this framework, 3 sets of experiments were conducted and the obtained results are reported in Tables~\ref{table: results experiment}, \ref{table: ocr results}. The goal of the $\mathbf{1}^{\mathrm{st}}$ experiment was to establish a baseline for images that would be classified as acceptable according to human standards. The appearance of unreadable images was especially prominent in the $1^{\text{st}}$ of the three datasets. Conversely, the average image quality of the $2^{\text{nd}}$ and $3^{\text{rd}}$ datasets was higher, therefore they were not considered in this experiment. Moreover, the $1^{\text{st}}$ dataset contained images from seven different locations, and as such, there were at least seven different types of food packaging present. To devise a balanced experiment, images from all locations were combined and categorised into $2$ classes: `Complete Dates' and `Unreadable'. 
As reported in Table~\ref{table: results experiment}, $\mathbf{90.1\%}$ classification accuracy was achieved over all seven locations.
\begin{table}[!t]
\caption{Number of images per category in each dataset.}
\label{table:dataset details}
\centering
\setlength{\extrarowheight}{2pt}
\setlength{\arrayrulewidth}{.4pt}
\begin{tabular}{c|c|c|c}
\hline \hline
\multirow{2}{*}{Annotation (DD/MM)} & \multicolumn{3}{c}{Dataset}\\ \cline{2-4}
& 1 & 2 & 3 \\ 
\hline
Missing/Missing & 375 & 3715 & 0 \\
Missing/Complete & 59 & 68 & 16  \\
Complete/Missing & 10 & 39 & 0  \\
Complete/Complete & 645 & 2847 & 1138 \\
Unreadable & 315 & 46 & 0 \\
\hline \hline
\end{tabular}
\end{table}
The $\mathbf{2}^{\mathrm{nd}}$ experiment aimed at distinguishing between acceptable and not-acceptable, missing dates. This meant that the absence of either day or month digits in a \textit{use-by} date is not acceptable. The $2^{\text{nd}}$ dataset was the largest, containing approximately $50\%$ of examples with partial or missing dates. Images missing the day/month or both were assigned to one class and `Complete Dates' to the other. As reported in Table~\ref{table: results experiment}, an accuracy of $\mathbf{96.8\%}$ was achieved. Similarly, a performance of $\mathbf{94.8\%}$ was achieved when applying the same procedure to the $1^{\text{st}}$ dataset. As for the $3^{\text{rd}}$ dataset, it includes images of higher quality, but there is a very small number of missing value examples available. To address this, we performed data augmentation in order to produce a larger set of `Partial Dates'. The accuracy achieved on this synthetic set was $\mathbf{85.8\%}$. Lastly, a small variation of this experiment (2.1 in Table~\ref{table: results experiment}) was conducted in order to assess how well the network can identify the presence of any type of date, be it complete or partial, versus the absence of a date altogether. This experiment offered insight into how well the network can produce inferred localisation of dates, as it must learn to filter out the abundant non-date related text/numbers in the images. Table~\ref{table: results experiment} shows that good accuracies were achieved across all three datasets, with the best case of $\mathbf{96.2\%}$ date presence detection on the $2^{\text{nd}}$ dataset. 

In a brief $\mathbf{3}^{\mathrm{rd}}$ experiment, a global approach to OCV was tested by targeting the classification of specific digits and letters. Successful text recognition systems typically begin with the detection of text presence within a given image, followed by a segmentation or localisation of the desired region-of-interest (ROI) in order to perform classification of segmented digits thereafter. Here we assess how well the NN can perform without specifying any additional labels or local information. Given that almost all images in the $3^{\text{rd}}$ dataset contained `Complete Dates', we conducted a brief digit classification experiment (see Table~\ref{table: ocr results} for results). Despite the small number of training examples ($1138$) and limited possible class combinations, four digit classes were identified, namely: $5$, $8$, $16$ and $20$. With these labelled examples, an accuracy of $\mathbf{90\%}$ was achieved. Similarly for the $2^{\text{nd}}$ dataset - due to limited data - a brief global OCV classification experiment between the the months of October and Novemeber in \textit{use-by} dates was conducted. An accuracy of $\mathbf{92.7\%}$ was achieved despite the small number of training examples.
In reflection of these results, it is important to remember the great variety of text and numbers included in each image. Without providing any local knowledge and given limited training examples, the networks were still able to automatically infer the importance of specific digits and their respective locations in a global manner, whilst ignoring the same or other digits located in close proximity. 
\subsubsection{Latent Variable Adaptive Clustering}
A major challenge spanning the three datasets was the high variability in the captured images characteristics. This variability made the reuse of a DNN trained on one dataset, for classifying the data of another, very difficult leading to poor performances. Fundamentally, this is because each dataset comes from a different distribution, as the images were taken by different people, with different cameras and at differing supplier locations. With limited data available to us, the use of transfer learning among different environments and datasets was ineffective. To overcome these challenges, we demonstrate the possibility of designing a new facet of the same CNN architecture, for learning each considered problem associated with different datasets. The approach focuses on: \textit{i)} detecting bad image capturing conditions; \textit{ii)} detecting missing dates (\textit{i.e.} either day and/or month of \textit{use-by} date); \textit{iii)} showing the ability to recognise day and/or month of an existing \textit{use-by} date. The CNN architectures proved to be quite accurate in identifying the missing/complete dates classification problem. Subsequently, we explored whether the respective trained networks were suitable for carrying out the proposed network adaptation approach (see Table~\ref{table:results_experiment_adaptation} for results). 
\begin{table}[!t]
\caption{Experiment results of OCV binary classification.}
\label{table: results experiment}
\centering
\setlength{\extrarowheight}{2pt}
\setlength{\arrayrulewidth}{.4pt}
\begin{tabular}{c|c|c|c|c}
\hline \hline
\multicolumn{5}{c}{CNN Optical Character Verification} \\
\hline
Exper. & Dataset & OK & NOT-OK & Accuracy ($\%$) \\
\hline
1 & 1 & 645 & 645 & \textbf{90.1}\% \\
\hline
\multirow{3}{*}{2} & 1 & 645 & 444 & 89.3\% \\ 
& 2 & 2847 & 2847 & \textbf{96.8}\% \\          
& 3 & 577 & 577 & 85.8\% \\ 
\hline
\multirow{3}{*}{2.1} & 1 & 714 & 375 & 94.8\% \\
& 2 & 2954 & 2954 & \textbf{96.2}\% \\ 
& 3 & 199 & 199 & 88.1\%  \\ 
\hline \hline
\end{tabular}
\end{table}
\begin{table}[!t]
\caption{Experiment results for date character recognition.}
\label{table: ocr results}
\centering
\setlength{\extrarowheight}{2pt}
\setlength{\arrayrulewidth}{.4pt}
\begin{tabular}{c|c|c|c}
\hline \hline
\multicolumn{4}{c}{CNN Date Character Recognition} \\
\hline
Exper. & Dataset & Images per Class & Accuracy ($\%$) \\
\hline
\multirow{2}{*}{3} & 2 & 381, 381, 381 & \textbf{92.7}\% \\
& 3 & 55, 67, 63, 61 & 90\% \\
\hline \hline
\end{tabular}
\end{table}

\begin{figure}[!b]
    \centering
    \includegraphics[width=.49\textwidth]{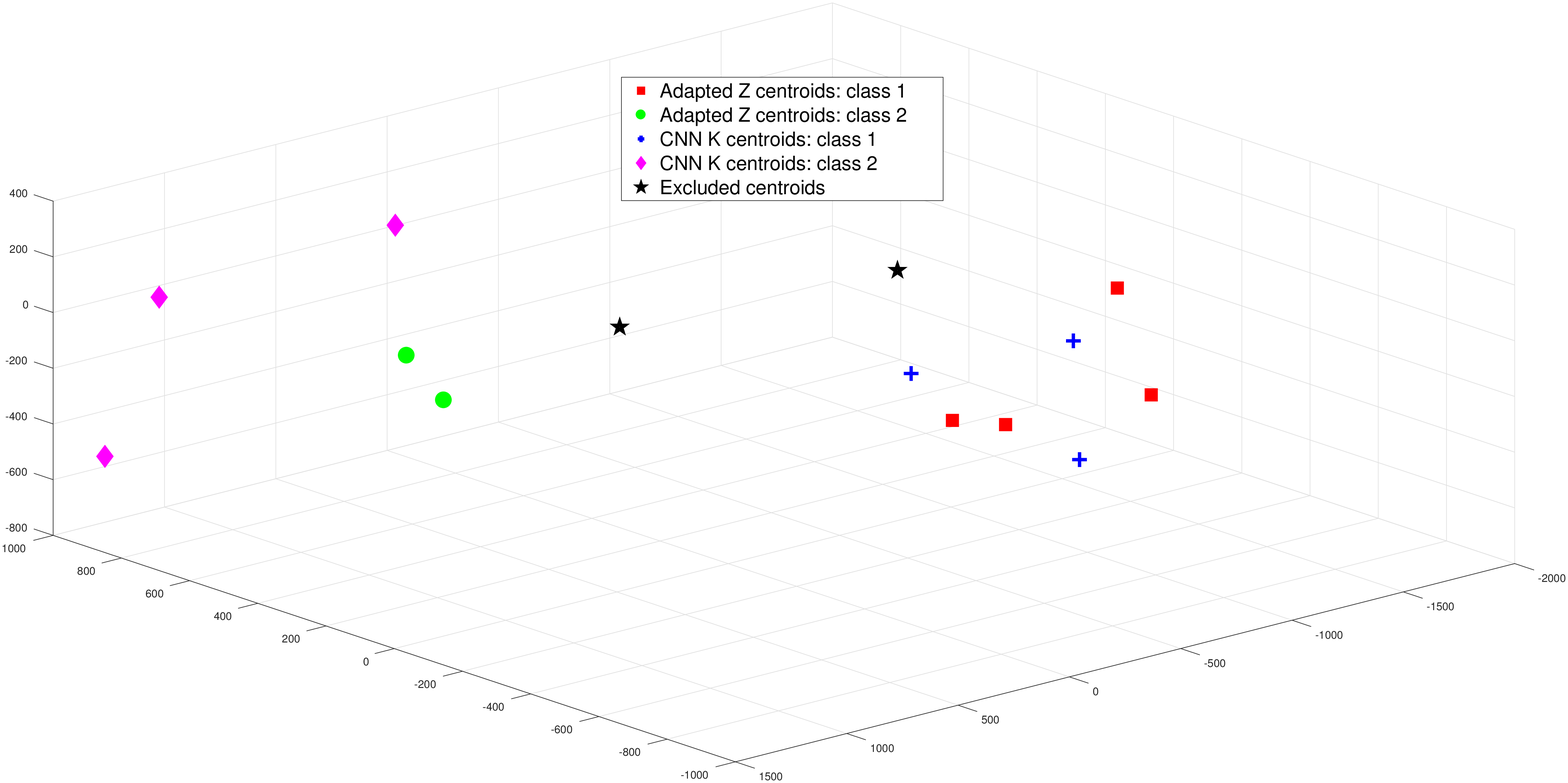} 
\caption{t-SNE visualisation of the derived centroids $\textbf{A}$ with best $k=7$, achieving the results reported in Table~\ref{table:results_experiment_adaptation}. The `Excluded centroids' ($2$ black stars) were removed as per the policy outlined in step 6 of our proposed adaptation procedure.}
\label{fig: cluster_centroids}
\end{figure}

To this end, consider $\mathcal{F}(\mathcal{D}_2; \mathbf{W}_2)$ as a trained CNN with a test performance of $95.9\%$ on a binary classification problem of \textit{use-by} date verification on a real dataset. Let $\mathcal{T}_{1}$ be the test set of a dataset from a different distribution targeting the same classification task. We forward-propagate $\mathcal{T}_{1}$ through $\mathcal{F}(\mathcal{D}_2; \mathbf{W}_2)$ and achieve a lower accuracy of $63.8\%$ as expected. We employed our adaptation procedure to classify $\mathcal{T}_{1}$ without any parameter retraining, decreasing the relative error by $34.81\%$ with an improved accuracy of $\mathbf{76.4\%}$. Interestingly, the original performance achieved by $\mathcal{F}(\mathcal{D}_2; \mathbf{W}_2)$ on $\mathcal{T}_2$ also increased from $95.9\%$ to $\mathbf{97.1\%}$ when classifying $\mathcal{T}_{2}$ with $\textbf{A}$ instead of the CNN it was originally trained on. Figure~\ref{fig: cluster_centroids} depicts a 3D visualisation of all $2048$-dimensional cluster centroids, for $k=7$ for both datasets ($14$ in total). Squares (Red) and (Blue) crosses denote the centroids corresponding to the complete date class in the first and second datasets respectively. (Green) circles and (Pink) diamonds are the centroids in the missing date category and the (Black) stars indicate the centroids not used in the final classification as per the centroid exclusion policy explained previously in section~\ref{subsec-Adaptation}.
\subsubsection{Deep Bayesian Self-Training on Real Data}
\begin{figure*}
    \centering
    \setlength{\tabcolsep}{0.3em}
    \begin{tabular}{cccc}
        \subfloat[]{\includegraphics[trim={0 0 0 0},clip,width=0.24\textwidth]{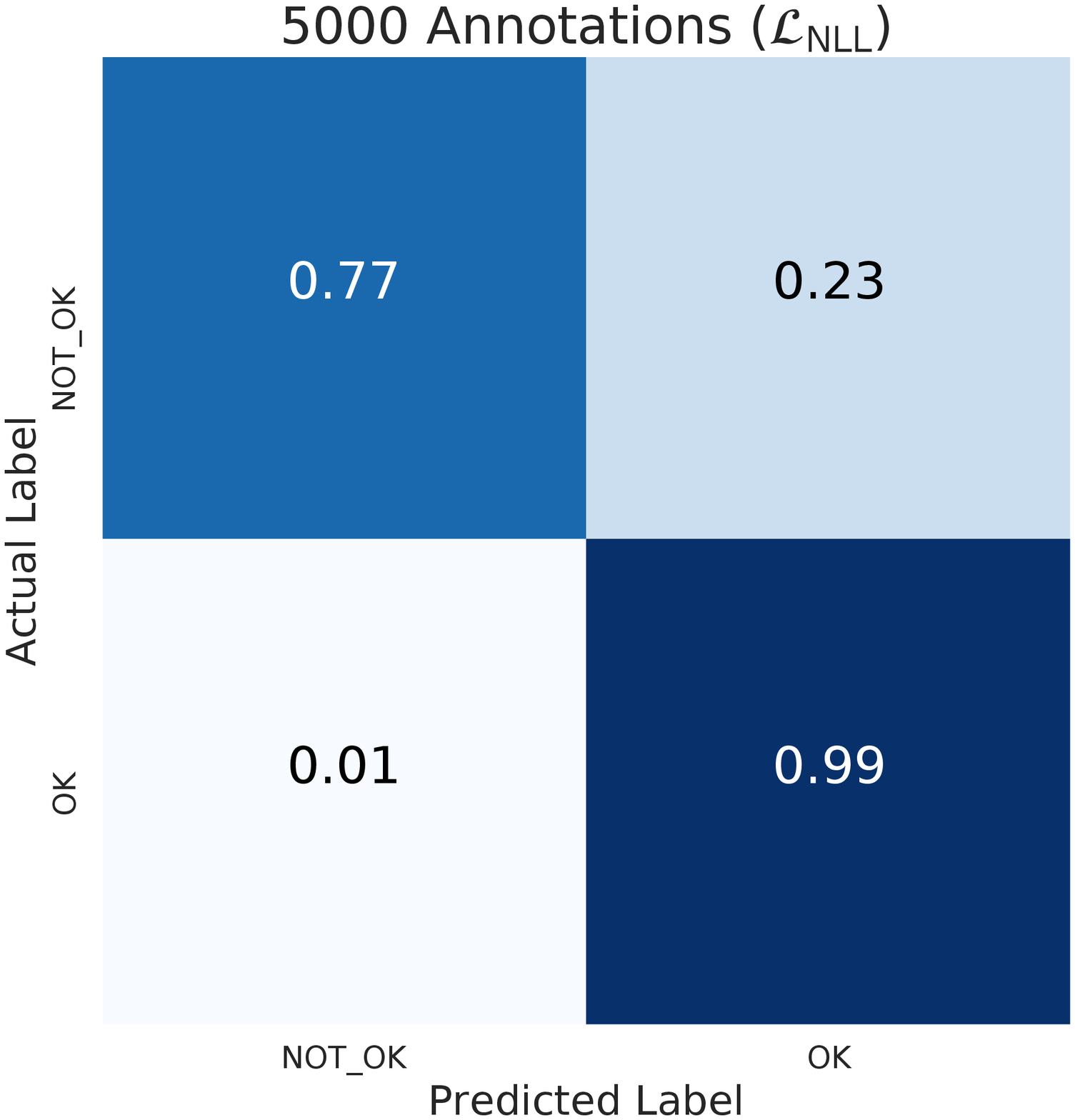}\label{conf_matrices-a}}&
        \subfloat[]{\includegraphics[trim={0 0 0 0},clip,width=0.24\textwidth]{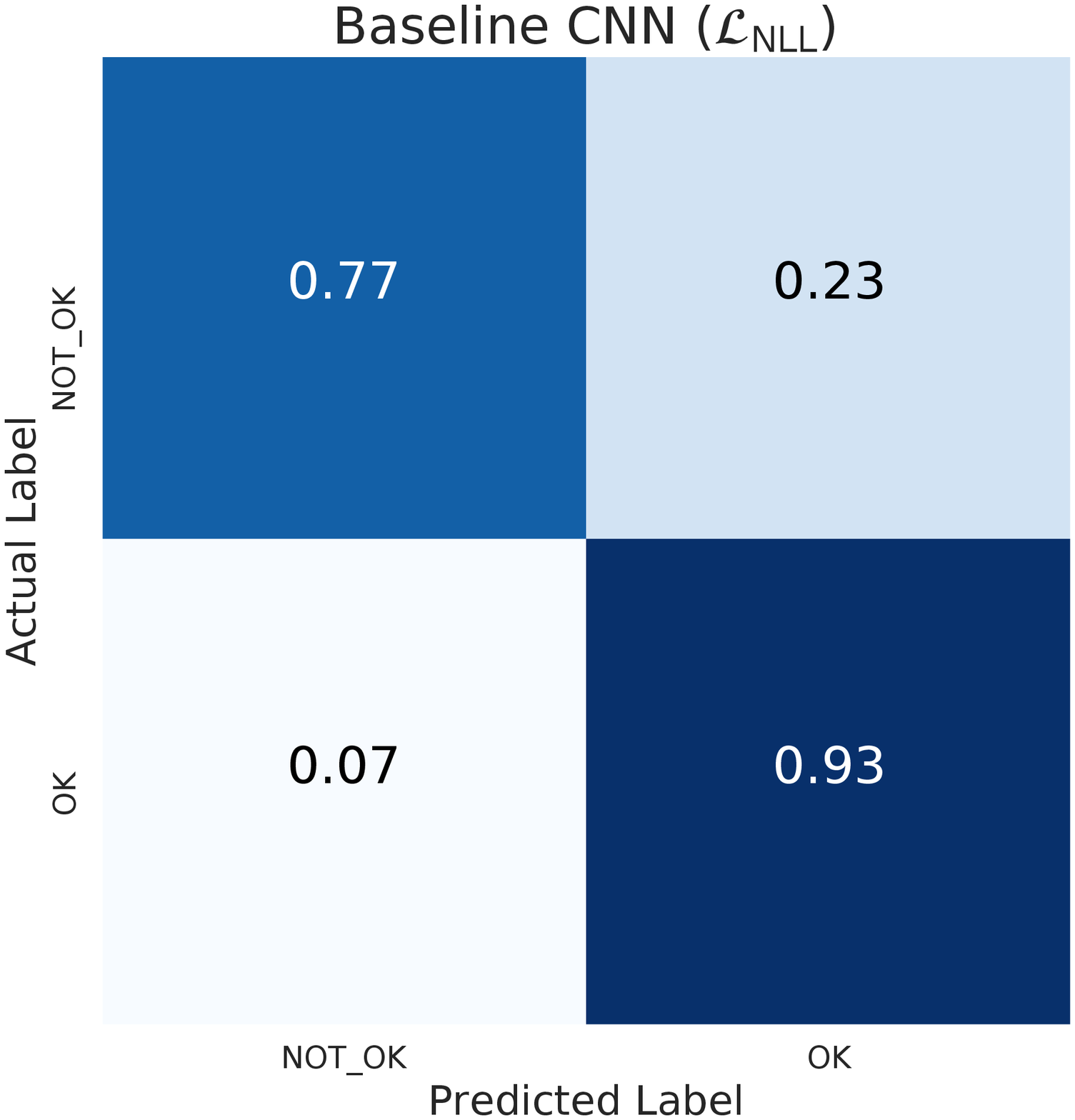}\label{conf_matrices-b}}&
        \subfloat[]{\includegraphics[trim={0 0 0 0},clip,width=0.24\textwidth]{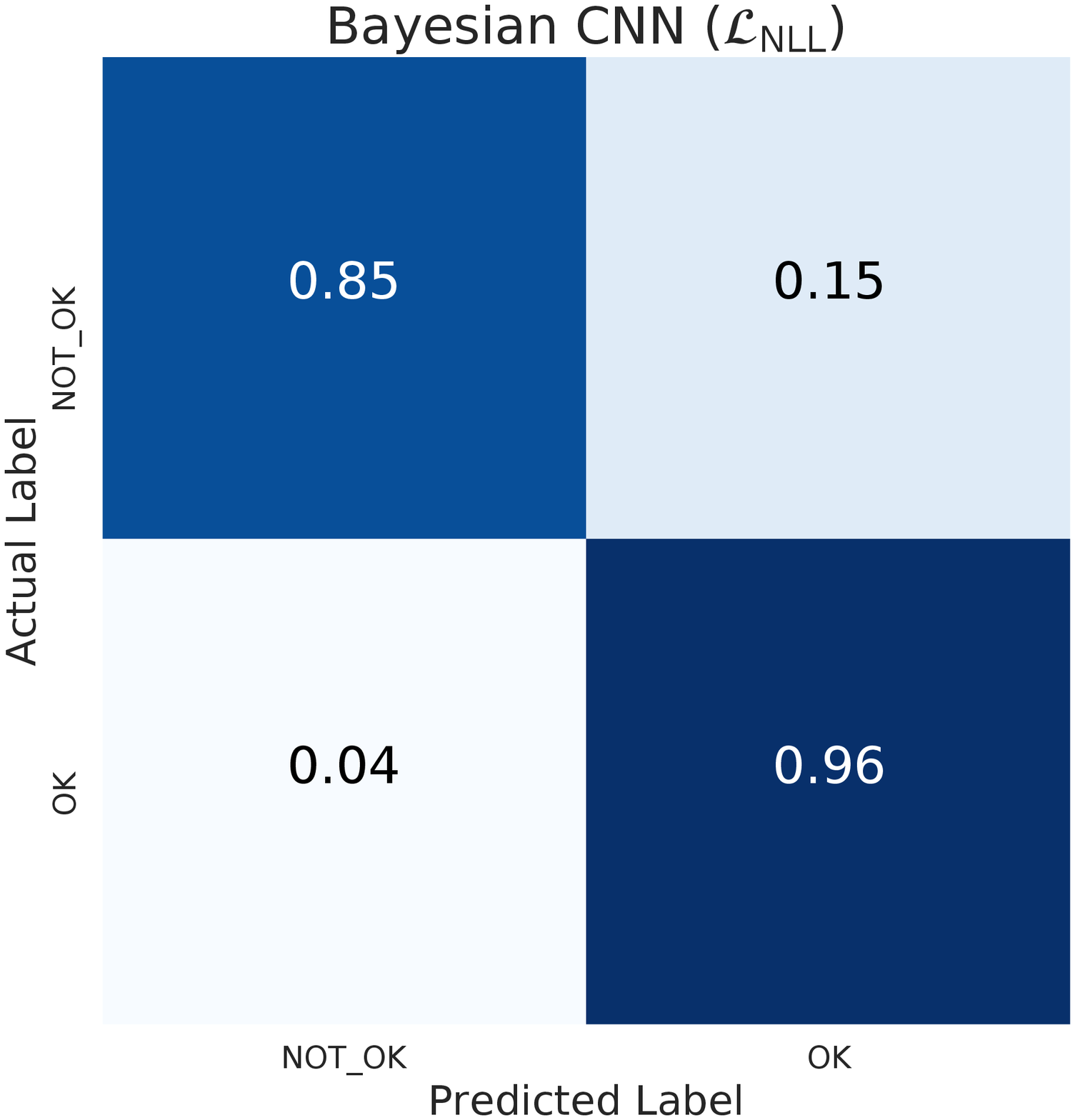}\label{conf_matrices-c}}& 
        \subfloat[]{\includegraphics[trim={0 0 0 0},clip,width=0.24\textwidth]{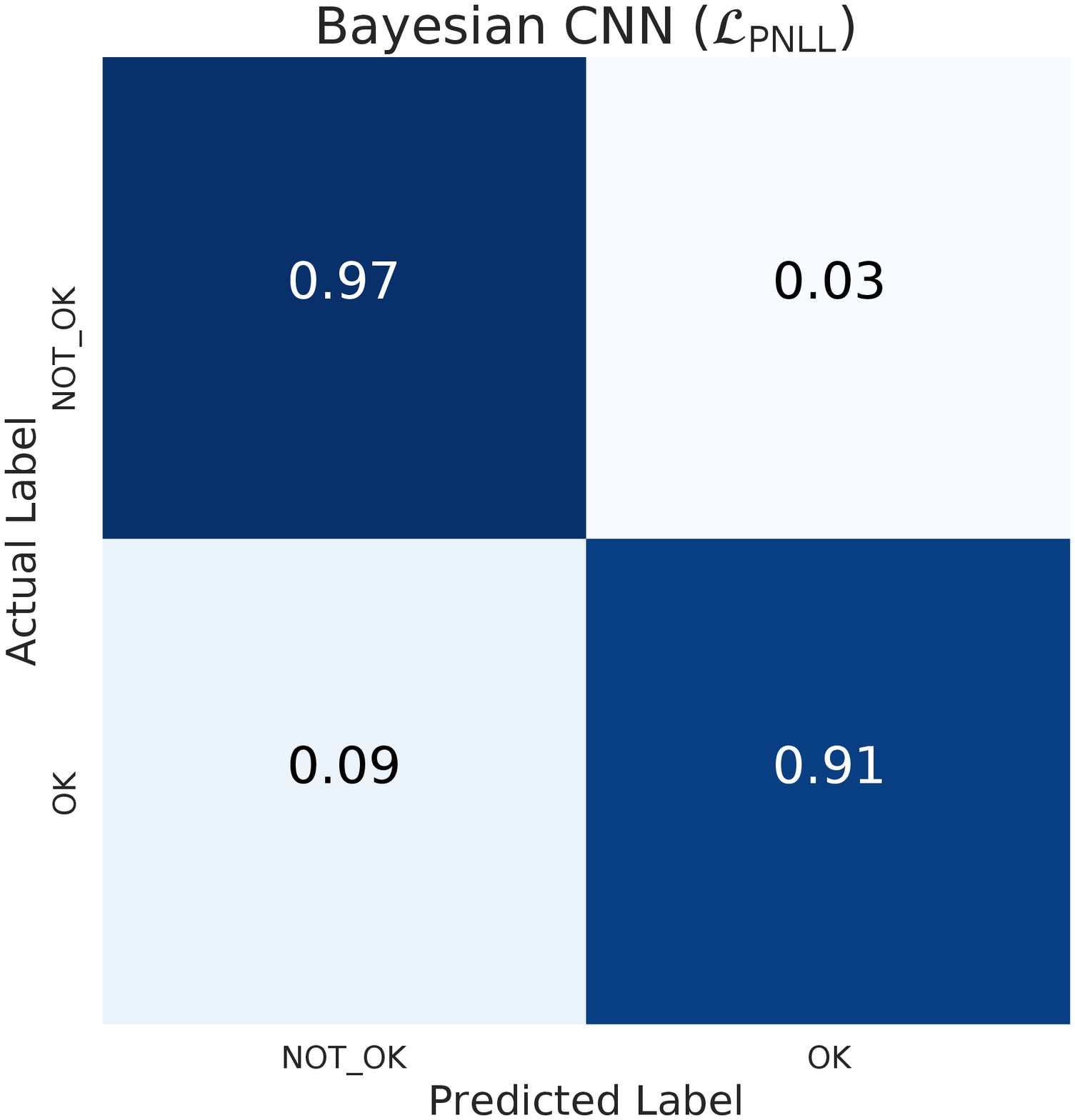}\label{conf_matrices-d}}
    \end{tabular}
 \caption{Normalised confusion matrices of the results obtained from our self-annotation procedure. 
 \textbf{(a)} Refers to the 5000 predicted labels obtained with the lowest prediction uncertainty. \textbf{(b)} Deterministic baseline CNN predicted labels, wherein the thresholds were set based on the network's sigmoid output. \textbf{(c)} Predicted labels from our Bayesian Self-Training approach, trained with a standard binary negative log-likelihood loss. \textbf{(d)} Similar to (c) but using a Bayesian CNN trained with the entropy penalised binary negative log-likelihood loss.}
\label{conf_matrices}
\end{figure*}

In order to validate our approach, we conducted a series of experiments on a pool of held-out annotated data comprised of $11948$ real food package images. The results can be seen in Table~\ref{tab:Self-annotating performance} and Figure~\ref{conf_matrices}. We begin by introducing Concrete Dropout layers after every convolutional layer in the last DenseBlock of a DenseNet-201, pre-trained on ImageNet. We then fine-tuned the last DenseBlock on a small portion of $500$ images, with binary annotated labels representing whether the \textit{use-by} date was readable (OK) or not (NOT-OK). 
\begin{table}[!t]
\caption{Experiment results of our adaptation procedure.}
\label{table:results_experiment_adaptation}
\centering
\setlength{\extrarowheight}{2pt}
\setlength{\arrayrulewidth}{.4pt}
\begin{tabular}{c|c|c}
\hline \hline
\multicolumn{3}{c}{Latent Variable Adaptive Clustering} \\
\hline
\multirow{2}{*}{Test Dataset} & \multicolumn{2}{c}{Classification Accuracy $(\%)$} \\ \cline{2-3} 
& CNN $\mathcal{F}(\mathcal{D}_2; \mathbf{W}_2)$ & Our Method ($\mathbf{A}$) \\ 
\hline
$\mathcal{T}_{1}$  &  63.8\% & \textbf{76.4}\%\\
$\mathcal{T}_{2}$  &  95.9\% & \textbf{97.1}\% \\
\hline \hline
\end{tabular}
\end{table}
As observable in Figure~\ref{conf_matrices-a}, we first applied these ideas to the full set of unlabelled $11948$ images and simply selected the $500$ most certain predicted labels to be added to the initial training set of $500$ images. This process was repeated $10$ times in order to collect a total of $5000$ images with predicted labels, which we then compared with our annotated labels as shown in Table~\ref{tab:Self-annotating performance}. In the remaining set of experiments, instead of selecting a pre-determined number of images, we filtered out uncertain predictions based on a threshold $\tau$ as in Algorithm 1. Figures~\ref{conf_matrices-c} and \ref{conf_matrices-d} depict the confusion matrices for the automatically annotated images w.r.t. true labels, and highlight the benefits of applying a confidence penalty on the log-likelihood loss ($\mathcal{L}_{\mathrm{PNLL}}$), as opposed to using a standard log-likelihood ($\mathcal{L}_{\mathrm{NLL}}$) which often outputs overconfident distributions. The uncertainties were calculated based on $50$ Monte Carlo Dropout samples at test time, following the description in section~\ref{subsec: inverse uncertainty weighting}. In order to compare our approach to standard Self-Training, we took the same network and datasets splits, and trained it without the Bayesian components. The threshold was set based on the confidence of the CNN output to only consider very confident predictions with over $0.999$ predicted probability. As can be seen in Table~\ref{tab:Self-annotating performance}, even with a high threshold, the deterministic CNN tends to be overconfident in its wrong predictions. This causes an increase of the propagated error as more images with wrong predicted labels are added to the training set and the model starts to under perform. To ensure a fair comparison between the Self-Training methods, the \textit{stop} conditions were set to be identical s.t. the procedure was interrupted after $3$ consecutive iterations without selecting more images to be added to the training set. 
\begin{table}
    \centering
    \setlength{\extrarowheight}{2pt}
    \setlength{\arrayrulewidth}{.4pt}
    \caption{Deep Bayesian Self-Training performance on real datasets. Cohen's Kappa score $\kappa$ is also reported.}
    \begin{tabular}{c|c|c|c|c}
    \hline \hline
    \multicolumn{5}{c}{Bayesian CNN ($\mathcal{L}_{\mathrm{PNLL}}), \ \kappa = \mathbf{0.8891}$} \\
     \hline
    Class & Precision & Recall & F1-score & \#img \\
    \hline
    NOT-OK & 0.9532 & 0.9694 & 0.9612 & 294 \\
    OK & 0.9427 & 0.9136 & 0.9279 & 162 \\    
    \hline
    Avg./Total & 0.9494 & 0.9496 & \textbf{0.9494} & \textbf{456} \\
    \hline \hline
    \multicolumn{5}{c}{} \\
    \hline \hline
    \multicolumn{5}{c}{Bayesian CNN ($\mathcal{L}_{\mathrm{NLL}}), \ \kappa = \mathbf{0.8383}$} \\
     \hline
    Class & Precision & Recall & F1-score & \#img\\
    \hline
    NOT-OK & 0.9679 & 0.8538 & 0.9073 & 212 \\
    OK & 0.889 & 0.9764 & 0.9306 & 254 \\ 
    \hline
    Avg./Total & 0.9248 & 0.9206 & \textbf{0.9200} & \textbf{466} \\
    \hline \hline
    \multicolumn{5}{c}{} \\
    \hline    \hline
    \multicolumn{5}{c}{Baseline CNN ($\mathcal{L}_{\mathrm{NLL}}), \ \kappa = \mathbf{0.6964}$} \\
     \hline
    Class & Precision & Recall & F1-score & \#img\\
    \hline
    NOT-OK & 0.9158 & 0.7682 & 0.8355 & 453 \\
    OK & 0.7989 & 0.9287 & 0.8589 & 449 \\
    \hline
    Avg./Total & 0.8576 & 0.8481 & \textbf{0.8472} & \textbf{902}\\
    \hline \hline
    \end{tabular}
    \label{tab:Self-annotating performance}
\end{table}

\section{Conclusion \& Future Work}
\label{sec: conclusion}
In this paper we propose a Deep Bayesian Self-Training methodology that leverages modern approximate variational inference in DNNs to estimate predictive uncertainty during a Self-Training setting. Both aleatoric and epistemic uncertainties of predicted pseudo-labels for unseen data are estimated, and the samples with the lowest predictive uncertainty (highest confidence) are added to the training set in an automated manner. We offer ways to mitigate the known problem of propagating errors in Self-Training by including: (i) an entropy penalty on the log likelihood loss to punish over confident output distributions and facilitate thresholding, and (ii) an adaptive sample-wise weight on the influence of predicted pseudo-labelled samples over gradient updates to be inversely proportional to their predictive uncertainty. Lastly, we propose a new simple methodology for visualising and analysing variability between two dataset distributions in DNNs, and attempt to adapt information from one problem to the other by clustering learnt latent variable representations in the context of our application domain. An experimental study on both public and private (real) datasets is presented demonstrating the increased performance of our algorithm over standard Self-Training baselines, and also highlighting the importance of predictive uncertainty estimates in safety-critical domains.

Our future work will extend the experimental study to large dataset, consisting of about half a million real food packaging images, and we intend to apply the presented DNN based methodologies for adaptation and self-annotation of this data.  
\begin{acknowledgements}
The authors would like to thank Mr. George Marandianos, Mrs. Mamatha Thota and Mr. Samuel Bond-Taylor for manually annotating datasets used in this study and of course the reviewers for their constructive feedback that helped to improve the manuscript. We would also like to thank Professor Luc Bidaut for enabling this collaboration. \\ \textbf{Funding:} The research presented in this paper was funded by Engineering and Physical Sciences Research Council (Reference number EP/R005524/1) and Innovate UK (Reference number 102908), in collaboration with the Olympus Automation Limited Company, for the project Automated Robotic Food Manufacturing System.\\
\textbf{Conflict of Interest:} The authors declare that they have no conflict of interest.
\end{acknowledgements}
\bibliographystyle{unsrt}
\bibliographystyle{spmpsci}      
\bibliography{bibfile}   
%
%
%
\end{document}